%% file: main.tex
\documentclass{article} 

\usepackage{microtype}
\usepackage{hyperref}
\usepackage{url}
\usepackage{booktabs}

\usepackage[utf8]{inputenc} 
\usepackage[T1]{fontenc}    
\usepackage{amsfonts}       
\usepackage{nicefrac}       
\usepackage{xcolor}         
\usepackage{graphicx}
\usepackage{subcaption}
\usepackage{multirow}
\usepackage{booktabs}   
\usepackage{ragged2e}   
\usepackage{geometry}
\usepackage{subcaption}
\usepackage{amsmath}
\usepackage{listings}

\usepackage{lineno}

\definecolor{darkblue}{rgb}{0, 0, 0.5}
\hypersetup{colorlinks=true, citecolor=darkblue, linkcolor=darkblue, urlcolor=darkblue}

\title{Learning a Canonical Basis of Human Preferences from Binary Ratings}

\author{%
  \hspace{-1em}Kailas Vodrahalli\thanks{Stanford University. Email: \texttt{kailasv@stanford.edu}}\hspace{-1em}
  \and Wei Wei\thanks{Accenture. Email: \texttt{wei.h.wei@accenture.com}}\hspace{-1em}
  \and James Zou\thanks{Stanford University. Email: \texttt{jamesz@stanford.edu}}\hspace{-1em}
}
\date{}

\begin{document}

\maketitle


\input{sections/abstract}
\input{sections/introduction}
\input{sections/related_works}
\input{sections/human_preference_embedding}
\input{sections/preference_summaries}

\input{sections/validation_studies}
\input{sections/applications_to_models}
\input{sections/discussion}

\bibliographystyle{plain}
\bibliography{references}

\newpage
\input{sections/appendix}

\end{document}

%% file: sections/abstract.tex
\begin{abstract}
Recent advances in generative AI have been driven by alignment techniques such as reinforcement learning from human feedback (RLHF).
RLHF and related techniques typically involve constructing a dataset of binary or ranked choice human preferences and subsequently fine-tuning models to align with these preferences. 
This paper shifts the focus to understanding the preferences encoded in such datasets and identifying common human preferences. 
We find that a small subset of 21 preference categories (selected from a set of nearly 5,000 distinct preferences) captures >89\% of preference variation across individuals.
This small set of preferences is analogous to a canonical basis of human preferences, similar to established findings that characterize human variation in psychology or facial recognition studies. 
Through both synthetic and empirical evaluations, we confirm that our low-rank, canonical set of human preferences generalizes across the entire dataset and within specific topics.
We further demonstrate our preference basis' utility in model evaluation, where our preference categories offer deeper insights into model alignment, and in model training, where we show that fine-tuning on preference-defined subsets successfully aligns the model accordingly.
\end{abstract}

%% file: sections/introduction.tex
\section{Introduction}\label{sec:introduction}

One of the major breakthroughs that has enabled generative AI in recent years is the use of alignment techniques such as human feedback reinforcement learning (RLHF) \cite{christiano2017deep}. RLHF and related techniques are broadly applicable and have seen use in alignment for general utility and helpfulness \cite{bai2022training, bai2022constitutional, rafailov2024direct, azar2024general, ethayarajh2024kto}, decreasing toxicity \cite{amini2024direct, lee2024mechanistic}, or a myriad of other combinations of specific preferences \cite{wang2023helpsteer, yang2024rewards}. 

The general paradigm of RLHF techniques involves several steps. The first step is to construct a dataset of human preferences, typically in the form of binary choices between two possible outcomes or, more generally, as a ranking across many such choices. For language models, for example, this may come in the form of two possible answers to a user's question, with an annotation for which answer the user prefers. Subsequent steps involve augmenting this first, often relatively small dataset, and finally fine-tuning the base model to better align its responses with the preferences encoded in the aforementioned human preference dataset.

In this paper, however, we focus our attention to the human preference dataset. Rather than treat the dataset as simply a means for aligning the generative model, we seek to understand \textit{what preferences are actually encoded in that dataset}, and, more broadly, \textit{what preferences humans typically have}.

We find that a relatively small subset of preferences encodes much of the variation between people. This result is not surprising, as these types of results have previously been observed. The seminal work of Turk and Pentland on eigenfaces found that a small set of canonical human faces can capture most physical variation across people \cite{turk1991eigenfaces}. It has also long been an object of study in psychology to group people by a small set of personality traits. There are many examples of such groupings ranging from the popular Myers-Briggs test \cite{myers1962myers}, to the more academically relevant Big Five \cite{costa1999five}, or to the clinically relevant Minnesota Multiphasic Personality Inventory \cite{hathaway1951minnesota}.

Our main contributions are as follows. \textbf{(1)} We develop a method for characterizing the preferences contained within human preference datasets. While we focus on data that contains binary preferences, the method generalizes to datasets with ranked preferences as well. \textbf{(2)} As a byproduct of this method, we also generate a richly annotated preference dataset that contains a hierarchical categorization of preferences as well as topic categorizations; we release this dataset as part of our results. \textbf{(3)} We discover a low-rank canonical set of human preferences. Despite significant variation in human preferences across topics, this set generalizes and works at both a dataset level and a topic level. \textbf{(4)} We validate the discovered preference sets using both synthetic and empirical methods. \textbf{(5)} And finally, we demonstrate the utility of our preference decomposition in evaluating and fine-tuning models to better align with individual users.

Our dataset and associated code is made available at \href{https://github.com/kailas-v/HumanPreferencesBasis}{https://github.com/kailas-v/basis-of-human-preferences}.

\begin{figure}[h]
    \centering
    \begin{subfigure}{0.9\textwidth}
        \includegraphics[width=\linewidth]{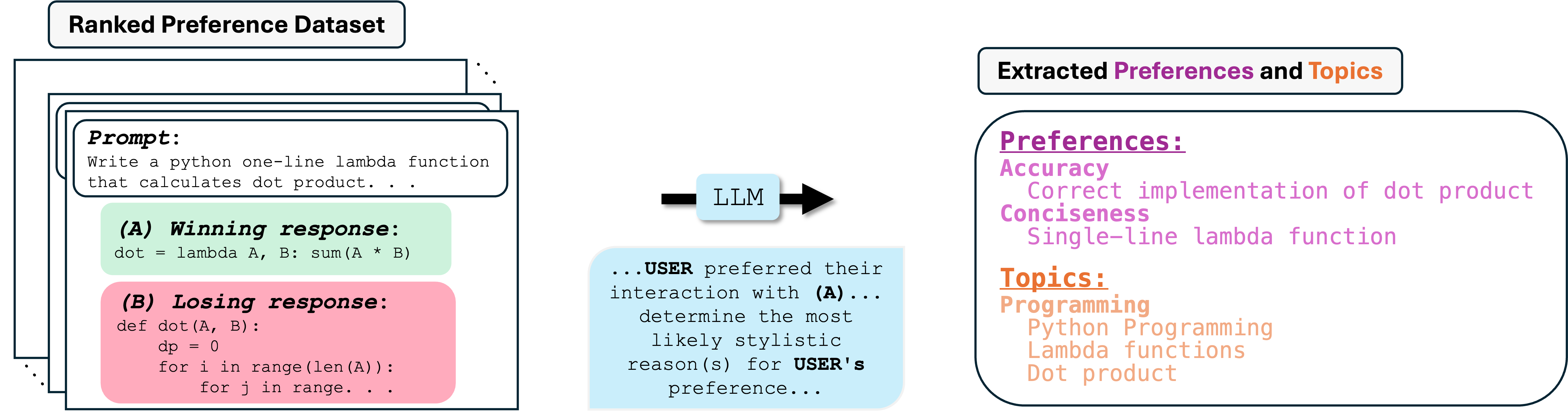}
        \caption{Each binary human choice is converted into a set of preferences and topic annotation.}
        \label{fig:overview_a} 
    \end{subfigure}
    \par\bigskip
    \begin{subfigure}{0.9\textwidth}
        \includegraphics[width=\linewidth]{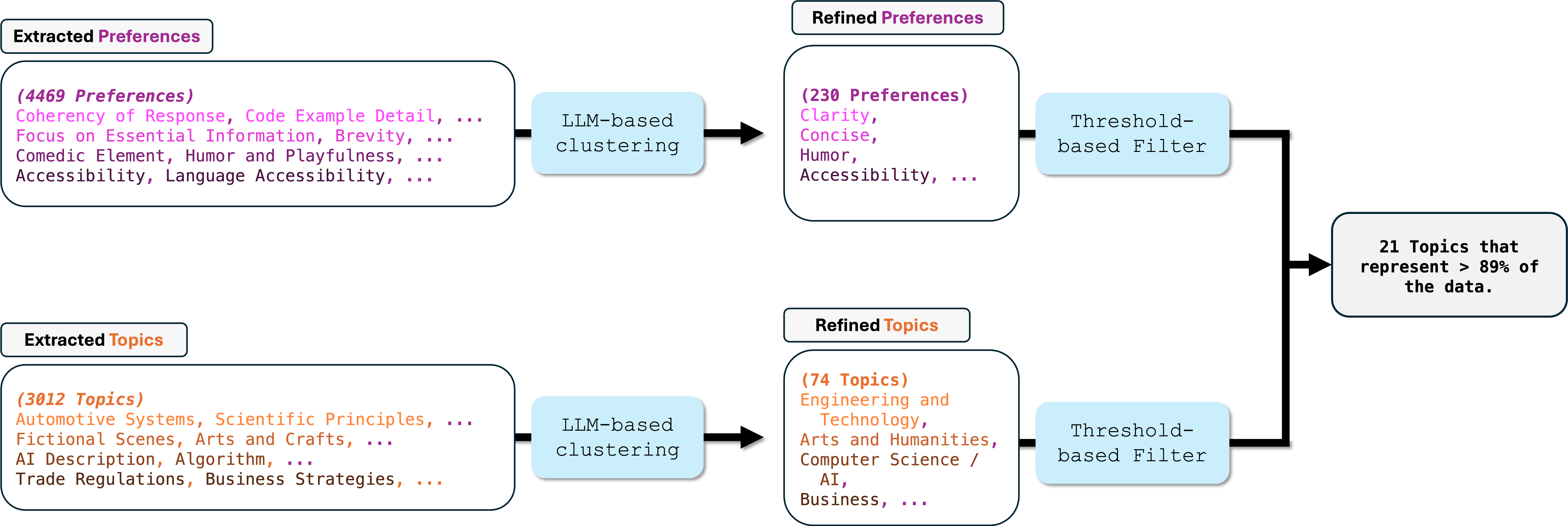} 
        \caption{Preferences and topics are aggregated and then independently refined, resulting in a small set of preferences covering most of the original dataset.}
        \label{fig:overview_b}
    \end{subfigure}
    \caption{Our pipeline converts a binary rating into a set of common human preferences. (A) This process is run in parallel for each binary choice. (B) This results in close to 5,000 preferences and over 3,000 topics. These preferences and topics are aggregated and then refined, resulting in just 21 preferences and 21 topics covering >89\% of the original dataset.}
    \label{fig:overview}
\end{figure}

%% file: sections/related_works.tex
\subsection{Related Works}\label{sec:related_works}

In the context of aligning large language models, human preferences are often encoded as binary or ranked choice data. The reason for this is that methods like RLHF and other competing methods typically involve training a reward model or optimizing directly using binary human preference comparisons \cite{christiano2017deep, rafailov2024direct, azar2024general}. These datasets can be categorized into two groups: (1) those that explicitly encode manually-selected preferences through either explicit annotations for the given preference(s) or through curated data collection \cite{bai2022training, bai2022constitutional, wang2023helpsteer}, and (2) those that focus on more generic helpfulness and utility data and not a specific set of preferences \cite{zheng2023judging, pmlr-v162-ethayarajh22a, h4stackexchange}. While some prior work catalogues the conversation topics present in these datasets \cite{zheng2023judging, chiang2024chatbot} or characterizes some of the preferences implicitly learned by the reward model \cite{singhal2023long, wang2024interpretable}, there has not been an explicit characterization of all the preferences encoded in generic preference data. Additionally, while the aforementioned datasets and methods are largely focused on improving model performance for a generic user, a few benchmarks and datasets instead focus on individual users, and so, encode a user-level set of preferences rather than high-level, aggregate user preferences \cite{salemi2023lamp, tan2024democratizing, poddar2024personalizing}. 

Our work is distinct in several ways. First, we seek to build on existing pairwise preference datasets by characterizing the preferences encoded in them. Secondly, by doing so, we aim to decompose the broad concepts of ``preference'' or ``helpfulness'' into something more interpretable. This characterization allows for not only a more transparent understanding of the specific preferences encoded in the data (e.g., ``concise'', ``humor'', or ``follows instructions''), but it is also useful for characterizing models and further aligning them along specific directions as dictated by individual users or task requirements.

%% file: sections/human_preference_embedding.tex
\section{Discovering a Representative Subset for Human Preferences}\label{sec:human_preferences}
To discover a canonical basis of human preferences, we leverage an existing dataset of binary human preferences. The choice to use binary preference data is not a limitation---ranked choice data can easily be adapted for our method by either converting it into a set of binary choice data or by slightly adjusting the model prompts we use. The binary choice dataset, which consists of human annotations for preferred responses, is used to uncover the implicit preference categories (e.g., ``likes concise responses'') that resulted in the binary choices. This set of discovered implicit preferences is then used to create a canonical model of human preference.

\subsection{Inferring Human Preferences from Binary Choice Data}\label{subsec:pipeline_step_1}
We start with an existing dataset of binary preferences. In particular, we use the Chatbot Arena dataset \cite{zheng2023judging}, which contains 33000 conversations with pairwise human preference ratings that compare the response of two different large language models. Using the two possible response options, we extract the \textit{reason} that best explains why a rater chose the preferred model response.

First, however, we filter the dataset for several criteria: (1) the language must be in English, (2) the human annotation cannot be a tie, and (3) we require single-turn conversations. Limiting to only single-turn conversations serves two purposes: it helps control where the preference comes from (i.e., the single response from each model rather than from somewhere else in the conversation) and it helps limit the conversation length. 
After filtering, we retain 18319 datapoints. 

Next, for each pair of AI responses, we query a large language model (LLM) to provide a list of preferences that can explain why a human preferred a given response. In our experiments, we use GPT-4o \cite{hurst2024gpt} for this task. The model is prompted to extract both a high-level preference as well as one or more more detailed descriptions of the preference. 
Here, the goal is to generate a short list of concise phrases that represent a cohesive preference. An examples is shown in Figure~\ref{fig:overview_a}, where the preference, ``conciseness'' is derived from an underlying, more detailed preference: ``Single-line lambda function.''
Note that a preference for ``conciseness'' does not alone encode directionality. A user may want more or less concise responses. We control for this by also prompting the model to write the preference from the perspective of wanting ``more'' of it.
We also simultaneously extract a list of one or more topics. 
Please see Appendix~\ref{app:llm_preference_prompts} for the prompts used and some additional examples.

\subsection{Refining Preferences to Derive a Canonical Subset}\label{subsec:preference_subset}
After running this query on all 18319 datapoints and performing basic string normalization, we retain 4469 unique preferences and 3012 unique topics. Many of these represent similar concepts, so we will further normalize the preferences using clustering. 

Using LLMs to cluster text has become common practice, with several methods showing improvements over baseline embedding-based clustering \cite{huang2024text, kwon2023image}. 
We take a similar approach. First, we prompt GPT-4o to generate a consistent labeling for each item in randomly sampled batches. This process performs best when the number of items is limited (i.e., less than 250), and so we repeat this process iteratively until all items have been clustered. We independently run this process for both preferences and topics, resulting in 230 preferences and 74 topics. 
Intuitively, these preferences and topics represent high-level categories of preferences and topics respectively.

Subsequently, we filter these preference and topic categories based on a simple threshold criteria: we keep preferences and topics that are present in at least 1\% of the dataset. This final filtering results in 21 preferences and 21 topics. The final count being equal across both preferences and topics is coincidental. This set of preferences represents $>89\%$ of all 4469 unique preferences.
The top-7 most common preferences and topics are shown in Tables~\ref{tbl:preferences} and Table~\ref{tbl:topics} respectively. All remaining topics and preferences are included in Appendix~\ref{app:all_preferences_and_topics}. 
While our work is the first to extract preferences from binary choice data, prior work has extracted conversation topic annotations. We find that our generated distribution of topics, which is skewed towards technical subjects, is similar to those found in prior works \cite{zheng2023judging, chiang2024chatbot}.

\begin{table}[t]
\begin{center}
\begin{tabular}{p{0.15\textwidth}p{0.1\textwidth}p{0.1\textwidth}p{0.25\textwidth}p{0.25\textwidth}}
\toprule
\textbf{Preference Category} & \textbf{\% of Data} & \textbf{\# of Preferences} & \textbf{Most prevalent in} & \textbf{Examples of granular preferences} \\
\midrule
\RaggedRight Clarity & 48.22\% & 474 & \RaggedRight Computer Science / AI, Engineering and Technology & \RaggedRight Situational Awareness, Contextual and Organizational Clarity \\
\midrule
\RaggedRight Thoroughness & 39.16\% & 414 & \RaggedRight Politics, Agriculture / Food & \RaggedRight Detail, Compositional Depth \\
\midrule
\RaggedRight Accuracy & 28.53\% & 248 & \RaggedRight Sports, History & \RaggedRight Precision, Accuracy in Context Application \\
\midrule
\RaggedRight Concise & 15.32\% & 28 & \RaggedRight General Knowledge, Sports & \RaggedRight Simplified Explanation, Simplicity of Language \\
\midrule
\RaggedRight Relevance & 15.13\% & 202 & \RaggedRight General Knowledge, Arts and Humanities & \RaggedRight Relevance to Query, Alignment with Game Themes \\
\midrule
\RaggedRight Engagement & 11.15\% & 237 & \RaggedRight Writing and Literature, Creativity / Innovation & \RaggedRight Engagement and Enthusiasm, Effective Hook \\
\midrule
\RaggedRight Innovation & 5.18\% & 96 & \RaggedRight Writing and Literature, Creativity / Innovation & \RaggedRight Originality, Creative Reasoning \\
\bottomrule
\end{tabular}
\end{center}
\caption{Most prevalent preference categories (selecting the top 7 by data percentage; all preferences are shown in Appendix~\ref{app:all_preferences_and_topics}). Note that datapoints may have multiple preferences; the majority have two preferences. So, the '\% of Data' column does not sum to 100. The third column is the count of preferences (from the original set of 4469 preferences) that cluster into this preference category.
}
\label{tbl:preferences}
\end{table}

\begin{table}[t]
\begin{center}
\begin{tabular}{p{0.3\textwidth}p{0.1\textwidth}p{0.1\textwidth}p{0.3\textwidth}}
\toprule
\textbf{Topic Category} & \textbf{\% of Data} & \textbf{\# of Topics} & \textbf{Most distinctive preference categories} \\
\midrule
\RaggedRight Engineering and Technology & 27.43\% & 304 & \RaggedRight Clarity, Accuracy \\ 
\midrule
\RaggedRight Arts and Humanities & 17.48\% & 368 & \RaggedRight Humor, Innovation \\ 
\midrule
\RaggedRight Computer Science / AI & 9.92\% & 179 & \RaggedRight Clarity, Direction \\ 
\midrule
\RaggedRight Business & 6.42\% & 270 & \RaggedRight Environment, Follows Instructions \\ 
\midrule
\RaggedRight Social Sciences & 4.73\% & 67 & \RaggedRight Customization, Diversity \\ 
\midrule
\RaggedRight Language and Communication & 4.66\% & 185 & \RaggedRight Environment, Innovation \\ 
\midrule
\RaggedRight Health & 3.50\% & 146 & \RaggedRight Direction, Helpfulness \\ 
\bottomrule
\end{tabular}
\end{center}
\caption{The final set of topics (selecting the top 7 by data percentage; all topics are shown in Appendix~\ref{app:all_preferences_and_topics}). The third column is the count of topics (from the original set of 3012 topics) that cluster into this topic category.
}
\label{tbl:topics}
\end{table}

%% file: sections/preference_summaries.tex
\section{Human Preference Archetypes}\label{sec:human_preference_summaries}
Here we provide a qualitative overview of the types of canonical preferences we uncover. 
While we find that only 21 preference categories are needed to cover most observed human preferences, we also find that both the distribution of preferences and the specific meaning of a preference depend on its topic and specific context.
This implies that while it is important to refine LLMs to be generally useful, understanding of the use case is also critical to ensure user alignment.

\subsection{Generic Preferences}
In general, there is a strong bias for clarity, thoroughness, accuracy, and conciseness across the data as indicated in Table~\ref{tbl:preferences}. This is partly due to the nature of the Chatbot Arena dataset. In the dataset, the AI models used span a wide range of performances. While the dataset includes more performant models like GPT-4, it also includes many smaller, less performant models. 
Additionally, there is a heavy bias in the dataset to technical subjects like computer science (see Table~\ref{tbl:topics}) and for general information requests that contribute to the bias for clear, accurate, and thorough information.

\subsection{Topic Specific Preferences}
\begin{figure}[h]
    \centering

    \begin{subfigure}{0.4\textwidth} 
        \includegraphics[width=\linewidth]{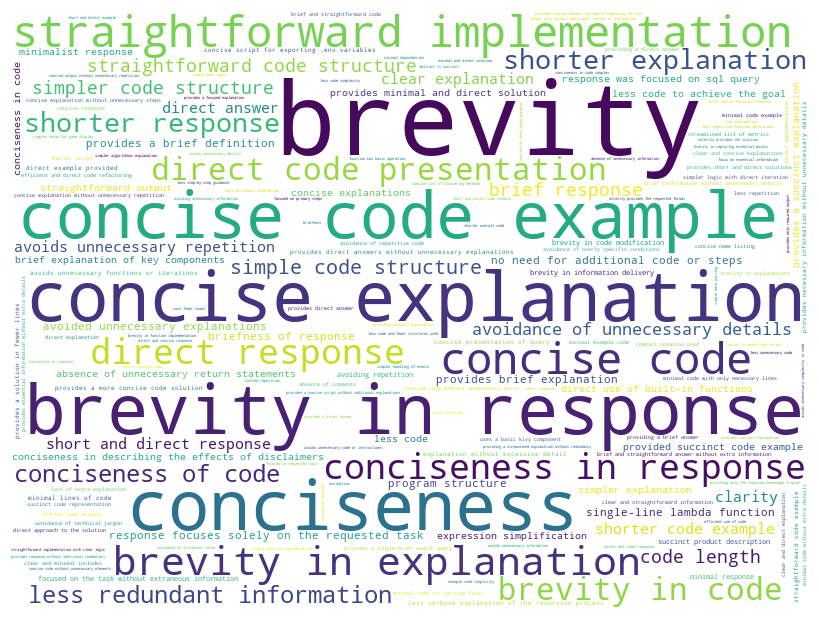} 
        \caption{Computer Science and AI, ``Concise''}
        \label{fig:wordclouds_a}
    \end{subfigure}
    \hfill 
    \begin{subfigure}{0.4\textwidth} 
        \includegraphics[width=\linewidth]{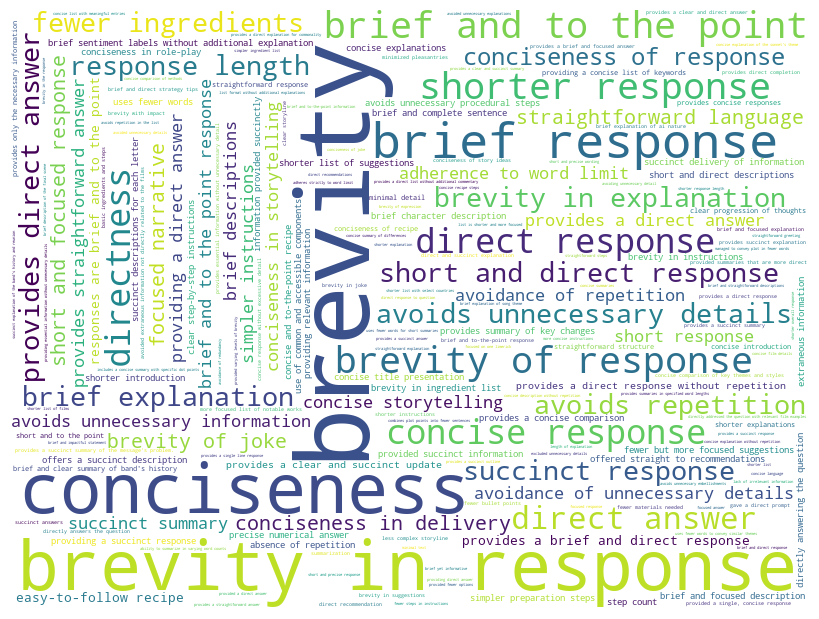} 
        \caption{Arts and Humanities, ``Concise''}
        \label{fig:wordclouds_b}
    \end{subfigure}

    \vspace{0.3cm} 

    \begin{subfigure}{0.4\textwidth} 
        \includegraphics[width=\linewidth]{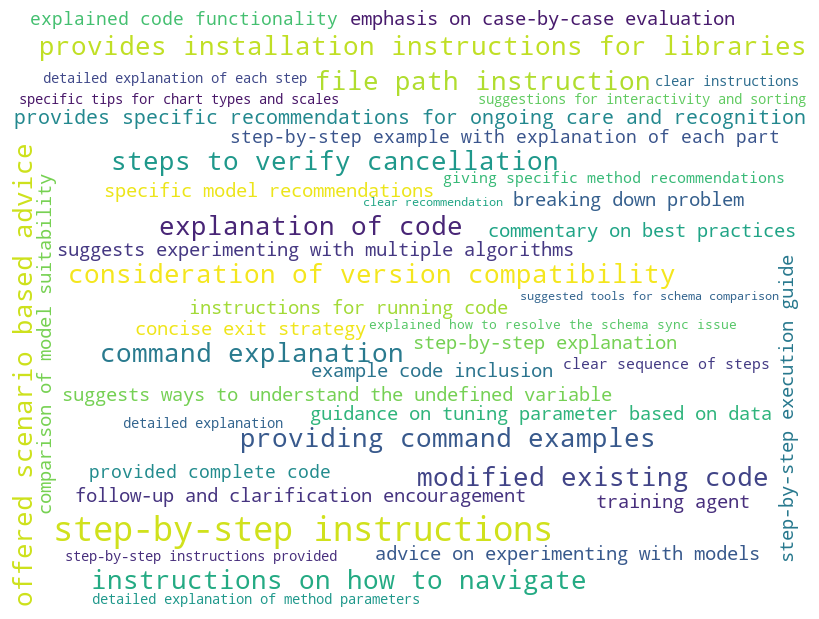} 
        \caption{Computer Science and AI, ``Direction''}
        \label{fig:wordclouds_c}
    \end{subfigure}
    \hfill 
    \begin{subfigure}{0.4\textwidth} 
        \includegraphics[width=\linewidth]{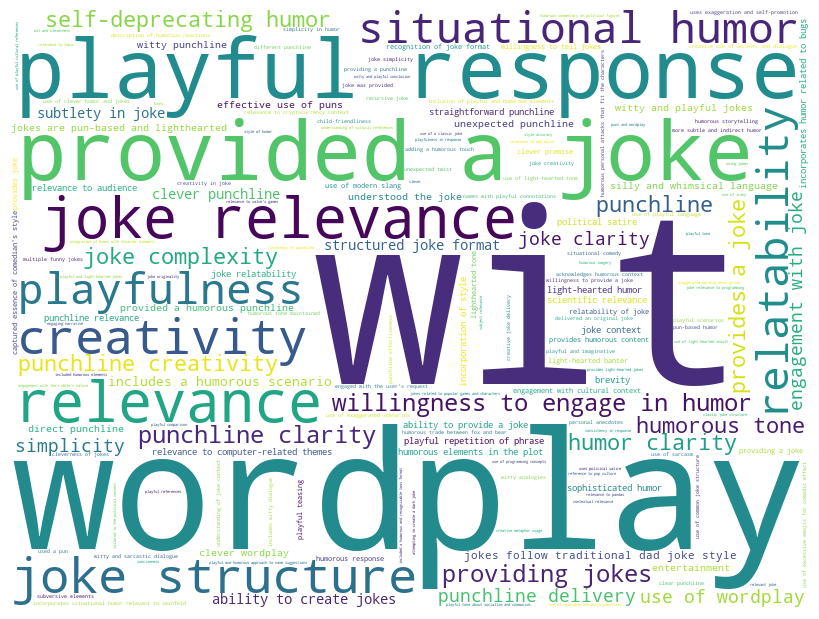} 
        \caption{Arts and Humanities, ``Humor''}
        \label{fig:wordclouds_d}
    \end{subfigure}
    \caption{Word clouds showing underlying, granular preferences.}
    \label{fig:wordclouds}
\end{figure}

In Figure~\ref{fig:wordclouds}, we show word clouds of the underlying preferences on which the preference categories are built. The top row shows two versions of ``Concise.'' While ``Concise'' has a similar meaning across topics, there are distinctions. For example, in Computer Science and AI, there is an emphasis on concise code descriptions and implementations, in addition to a more generic preference for short responses from the LLM.

We also find significant variation in the distribution of preferences across topics. This is illustrated in Figure~\ref{fig:preference_distribution} in the Appendix. For example, we observe that when users ask questions related to ``Computer Science and AI,'' they are most concerned with accuracy, clarity, thoroughness, and conciseness. Preferences for humor and engagement are almost nonexistent. 
In contrast, when conversations are related to ``Arts and Humanities,'' users care more about traits like engagement and innovation (i.e., ``creativity''), as well as a number of other traits like humor and diversity (e.g., ``considering multiple viewpoints''). While accuracy, clarity, thoroughness, and conciseness are still valued by users, they occur at below the average rate across all topics. 

%% file: sections/validation_studies.tex
\section{Evaluations}\label{sec:evaluation}
To evaluate our subset of preferences, we construct a multiple-multiple choice (MMC) benchmark, which allows selecting multiple answers in a multiple choice setting. We choose to use MMC questions because they are easy to give to both humans and LLMs (for synthetic evaluations). We find significant adherence to the LLM-extracted preferences and widespread agreement across three LLM evaluators and a cohort of human annotators.

\subsection{Evaluation methodology}\label{subsec:eval_methods}
First, some notation: let $d_i \in D$ be a pairwise comparison from the original binary preference dataset, $D$ (Chatbot Arena in our case). Let $p_{i,j}$ and $t_i$ be a preference and the topic category ascribed to $d_i$ respectively (recall that $d_i$ may have multiple preference categories but only one topic category; we randomly sample one preference here). Finally, define the following sets of \textit{granular} preferences:

\begin{itemize}
    \item $G_1 := G_{d_i, p_{i,j}}$: the set of granular preferences generated for $d_i$ and ascribed to preference $p_{i,j}$
    \item $G_2 := G_{p_{i,j}, t_i}$: the set of all granular preferences generated for any $d_j \in D$ with preference $p_{i,j}$ and topic $t_i$
    \item $G_3 := G_{t_i}$: the set of granular preferences generated for any $d_j \in D$ with topic $t_i$ 
    \item $G_4 := G_{p_{i,j}}$: the set of all granular preferences generated for any $d_j \in D$ with preference $p_{i,j}$
    \item $G_5 := G_\text{all}$: the set of all granular preferences
\end{itemize}

Now, we generate 6 choices, $C_1,\dots,C_6$, which are sampled as follows: for $i < 6$, $C_i \in G_i \setminus \bigcap_{j<i} G_j$ and $C_6 := \text{other reason(s)}$.

To assess our preferences, we compute four measures on the responses. Let $R_i$ be the fraction of responses selecting the choice from $G_i$. We compute four probability ratios: ``Generated vs. Control'' ($R_1/R_5$) compares generated preferences to random preferences; ``Generated vs. Control | Topic'' ($R_1/R_3$) compares generated preferences to random preferences within a topic; ``Category vs. Control'' ($R_4/R_5$) compares preferences within a category to random preferences; and ``Category vs. Control | Topic'' ($R_2/R_3$) compares preferences within a category and topic to random preferences within just the topic.

\subsection{Human Evaluations}\label{subsec:human_evaluation}
A cohort of 50 human raters were selected online through the Prolific website \cite{prolific}. Each human rater was given a set of 20 tasks. For each task, raters are shown a question and the two possible responses from the Chatbot Arena dataset. They are instructed that a separate group of humans has selected a response (the preferred response in the dataset) and are asked to assess why that response was preferred. The order and selection of questions is randomized across users. More information and an example showing the survey instructions and UI is shown in Appendix~\ref{app:survey_ui}.

\subsection{Evaluation Results}

\begin{figure}[h]
    \centering
    \includegraphics[width=5.5in]{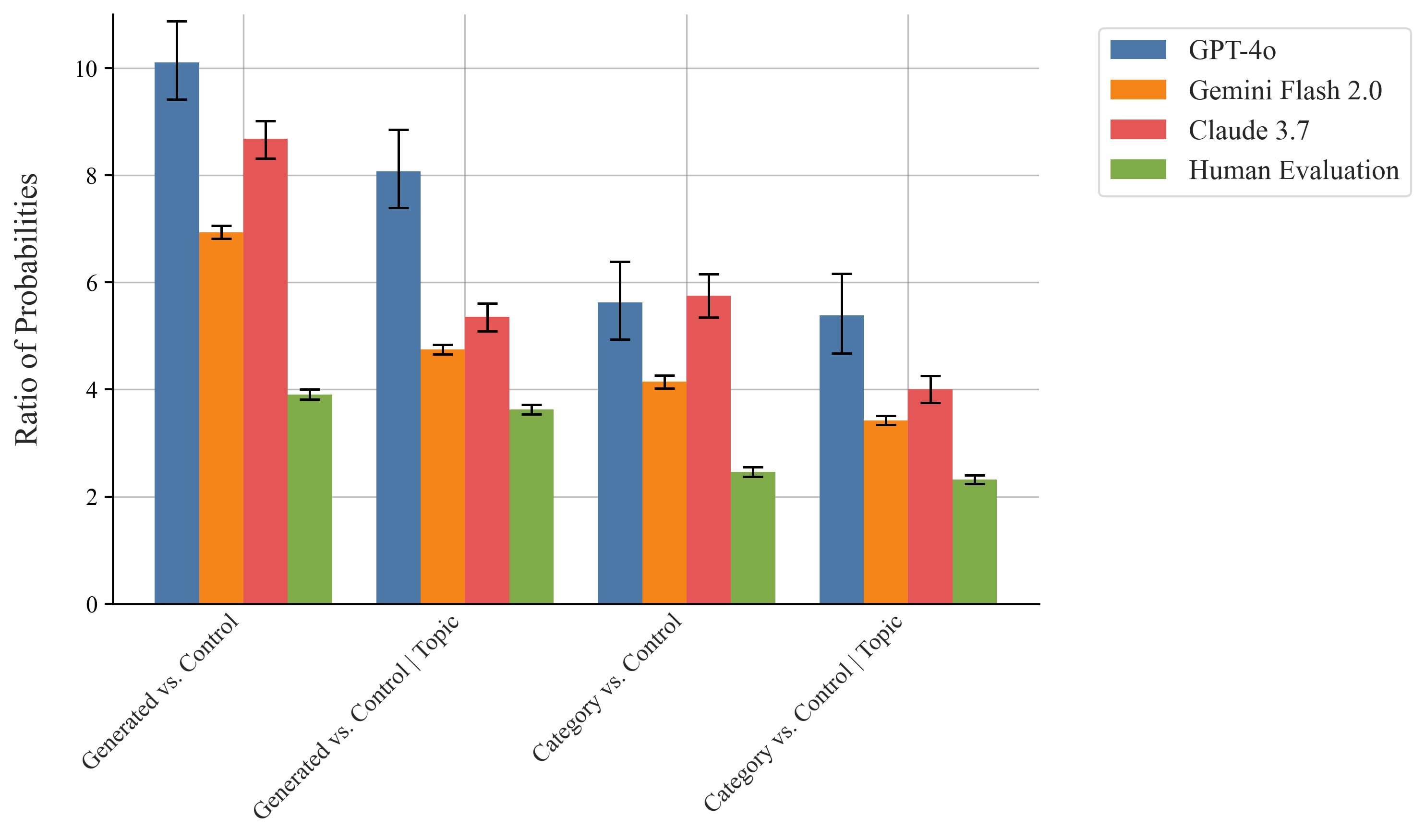}
    \caption{Probability ratios as described in \ref{subsec:eval_methods}. Comparison using GPT-4o \cite{hurst2024gpt}, Gemini \cite{team2023gemini}, and Claude 3.7 \cite{anthropic_claude_3_7_sonnet_2025}. A ratio of 1 would indicate no preference for the generated or category-specific preference. A ratio >~1 indicates preference for the generated or category-specific preference.}
    \label{fig:probability_ratios}
\end{figure}

In Figure~\ref{fig:probability_ratios}, we plot the four metrics previously described. We find that for all four metrics and across all three LLMs and the human evaluators, there is a significant difference above the baseline ratio (1, which would indicate no bias towards our granular preferences). 
In particular, ``Generated vs. Control'' indicates that the generated preference is much more strongly preferred to control preferences. This is still true when conditioning on a topic (``Generated vs. Control | Topic''), indicating that the preference categories we discovered are generalizable. That is, the same preference categories are useful for segmenting preferences in every topic. This finding, taken together with the often sparse distribution of preferences within a topic (see Figure~\ref{fig:preference_distribution} in the Appendix), indicates that conditioned on a given topic, even fewer preference categories may be sufficient to describe the majority of human variation.
The results for ``Category vs. Control'' and ``Category vs. Control | Topic'' both indicate that preferences sampled from categories are actually quite general. While using the actual generated topic for the specific example results in a higher rate of selection (i.e., more likely to be the underlying preference), sampling a random detailed preference from the preference category results in a significantly higher rate of selection than control. This result also generalizes across topics.

Additionally, the absolute rate of selecting the generated preference, $R_1$, is similarly high across all models ($>90\%$) and humans ($>70\%$). Finally, $R_6$, the rate that ``other reason(s)'' is selected, is $8.35\%$ in our human evaluations, indicating that the 21 preference categories together cover $>90\%$ of all real human preferences. More details are included in Appendix~\ref{app:additional_results}. 

%% file: sections/applications_to_models.tex
\section{Applications to Model Evaluation and Training}\label{sec:applications_to_models}

\subsection{Model Characterization}
We adapt the Elo ranking methodology used for arena-style leaderboards like in \cite{chiang2024chatbot} to generate preference-specific Elo (pElo) scores. For a given preference, its pElo score is computed by applying the Elo ranking algorithm solely to the subset of data labeled with that preference. Note that while GPT-4 is the best model overall, that is not true across all preferences. For example, GPT-3.5 (first) outperforms GPT-4 (third) in conciseness. The variation in preference is even more striking for other models like Palm 2, which has an overall rank of six but comes in fourteenth place for conciseness. More details and full rankings are included in Appendix~\ref{app:model_finetuning_and_eval}.

This characterization is important as it underscores the complex nature of model alignment. While generic alignment criteria like ``helpfulness'' are useful, the more fine-grained preferences we uncover allows us to move beyond generic evaluations and towards a more precise understanding of model strengths and weaknesses. Moreover, it is easy to apply pElo to existing leaderboard rankings. Our pipeline can be used to annotate binary preference data, which can subsequently be used to compute pElo rankings.

\subsection{Fine-tuning for Preference Alignment}\label{subsec:lora_finetuning}
We also find that fine-tuning on preference-defined subsets of data aligns the model with the given preference. We fine-tuned instruction fine-tuned versions of two models: Qwen2 7B and Ministral 8B \cite{qwen2, mistralai_ministral8b_instruct_hf_2024}. For each preference, we fine-tuned each model using Low-Rank Adaptation (LoRA) with Direct Preference Optimization (DPO) \cite{hu2022lora, rafailov2024direct}. Models were then evaluated on a held-out test set using an LLM-as-a-Judge setup adapted from \cite{zheng2023judging}. We find that fine-tuning results in a significant improvement in performance for nearly $40\%$ of preferences. This result is most striking when fine-tuning for ``conciseness,'' where we directly measure a $60\%$ reduction in response length. More details about the training and evaluation procedure as well as the complete fine-tuning results are included in Appendix~\ref{app:model_finetuning_and_eval}.

%% file: sections/discussion.tex
\section{Discussion}\label{sec:discussion}
In this work, we developed a pipeline to extract fine-grained preferences from binary preference data from which we identified a small, canonical set of preferences. We also validated these findings using simulation and empirical methods. Furthermore, we demonstrated the utility of this canonical set of preferences for evaluating and fine-tuning models for further alignment.

Future work can build on this foundation. 
As we demonstrated, we can fine-tune models along each of the preference directions. This can be the basis for individual-level (or task-level) personalization, where each user can be modeled by a linear (or non-linear) combination of these preferences; thus, a user may be characterized by their \textit{preference basis}, enabling rapid alignment of models to new users. 

As the use cases of LLMs grow more complex and nuanced, personalization becomes increasingly important. A general purpose LLM trained on generic preferences may not meet the personalized needs of a given user. Our work seeks to bridge this gap by identifying both at a high-level and in a more fine-grained way the preferences individuals care about. 

%% file: sections/appendix.tex
\appendix
\section{Dataset Information}\label{app:dataset_release}
We release the resulting preference annotations as well as the code used to generate them at \href{https://github.com/kailas-v/HumanPreferencesBasis}{https://github.com/kailas-v/basis-of-human-preferences}. 
The dataset here includes the outputs of each step. For each of the 18319 datapoints we release (1) the original set of preferences and topics generated using GPT-4o (totaling 4469 preferences and 3012 topics), (2) the detailed preferences underlying each of the generated preferences (e.g., see Figure~\ref{fig:wordclouds}), and (3) the preferences and topics after refinement along with their clustered preferences and topics respectively. Additionally, we release the analysis code used in Sections~\ref{sec:human_preference_summaries} and \ref{sec:evaluation}.

\section{Prompts to Generate Preferences}\label{app:llm_preference_prompts}
\begin{figure}[h]
    \centering
    \includegraphics[width=4.5in]{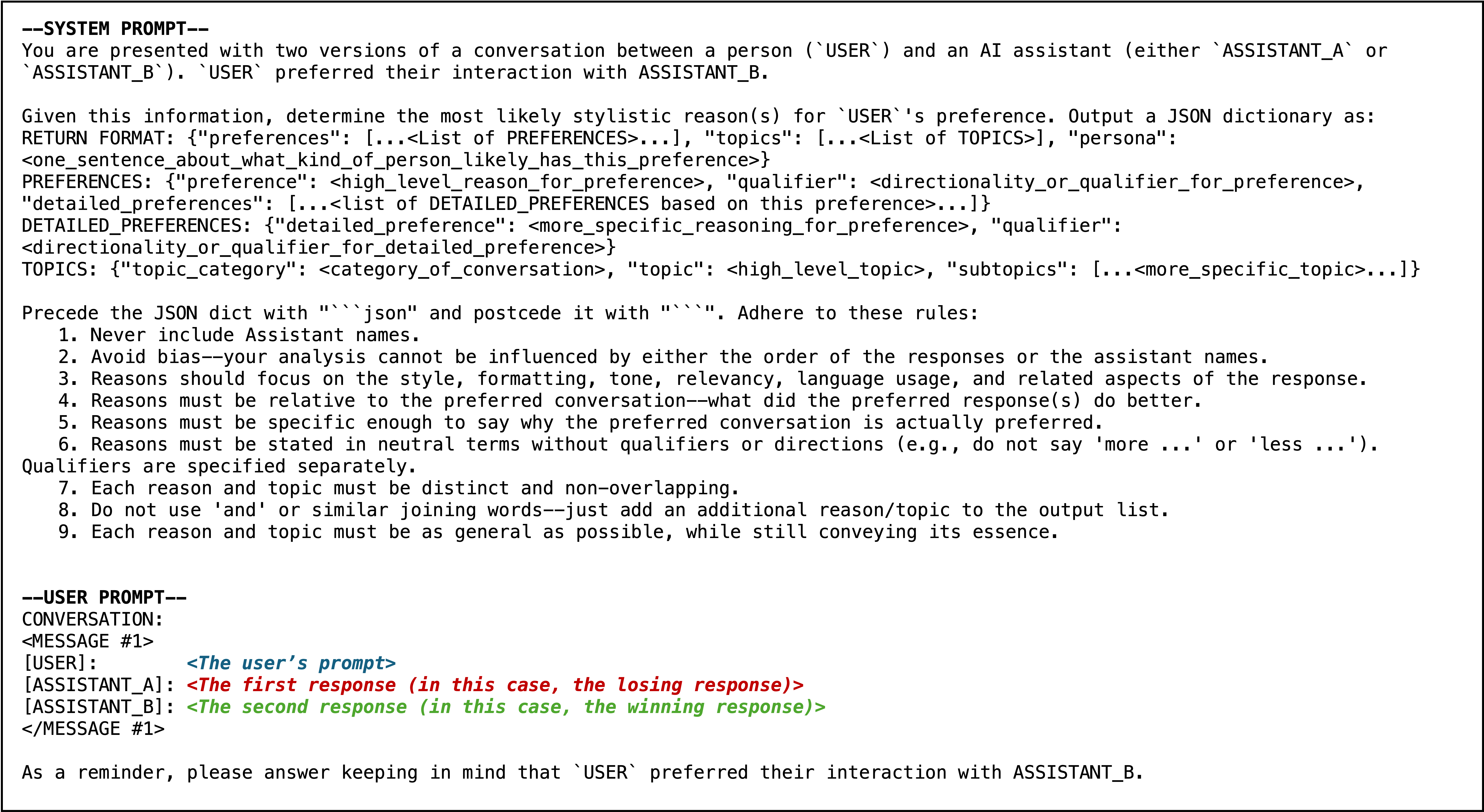}
    \caption{Prompt for extracting preferences and topics from binary comparison data.}
    \label{fig:extract_prompt}
\end{figure}

Preferences and topics are extracted in a multi-step process. The process is described below:

For each given conversation pair, the following process is run for preference and topic extraction (see Figure~\ref{fig:extract_prompt}):
\begin{itemize}
    \item The model is presented with the user's question along with the two candidate responses. The user's choice is provided to the model. We found that framing the prompt to ask the model for the reason for the user's choice, rather than the choice itself, lead to better performance.
    \item More specifically, the model is prompted to generate (A) a list of preferences along with (B) a list of topics, and (C) a short description of a persona for a user who might make the given choice. For each preference, and each topic, the model is also required to generate a list of more granular preferences and topics, respectively.
    \item We keep both the preferences and granular preferences, but found it more useful to only keep the high-level topics. We do not use the personas in our analysis, but release them with our dataset.
\end{itemize}

\begin{figure}[h]
    \centering
    \includegraphics[width=4.5in]{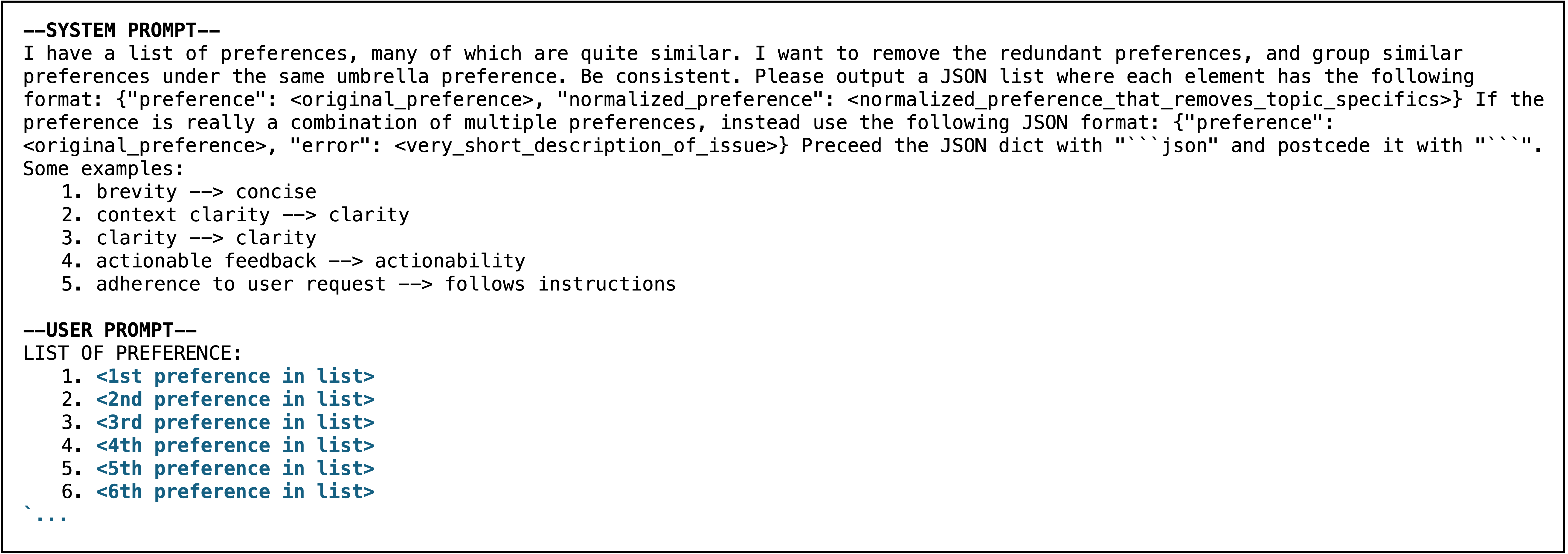}
    \caption{Prompt for refining preferences.}
    \label{fig:refine_prompt}
\end{figure}

After running this process for all conversation pairs, we are left with a list of preferences, granular preferences, and topics. We then refine the preferences and topics (see Figure~\ref{fig:refine_prompt}) to arrive at the canonical basis of 21 preferences and 21 high-level topics.


\section{Preferences and Topics}\label{app:all_preferences_and_topics}
All preferences are shown across Table~\ref{tbl:preferences_all_a} and Table~\ref{tbl:preferences_all_b} (split across two tables because of page space limitations). Together, the tables include the 21 preference categories, the percent of datapoints where the preference is present, details on how many preferences are contained in the category (from the original set of 4469 preferences), as well as the topics where the preference has the highest percentage representation and some examples of the granular preferences. All topics are shown in  Table~\ref{tbl:topics_all}. The topic table is similar to the Preference table, with the exception that we show here the most distinctive preference categories. These are the preference categories that are most overrepresented in the topic compared to their baseline representation. For example, ``Humor'' is very much overrepresented in ``Arts and Humanities'' as can be seen in Figure~\ref{fig:preference_distribution_b}; however, other preferences like ``accuracy'' and ``clarity'' are, on an absolute scale, more prevalent. We show this column as it is more informative on the variation across topics and suppresses the mean preferences of a generic person for accurate and clear responses.

\begin{figure}[h]
    \centering
    \includegraphics[width=4.5in]{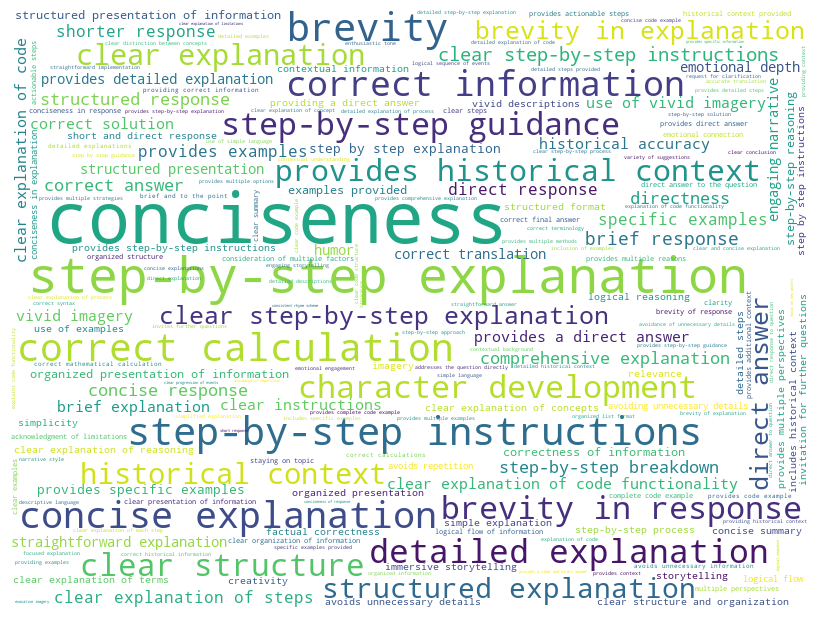}
    \caption{Word cloud of the preferences people have across all conversation discussion topics.}
    \label{fig:wordcloud_all}
\end{figure}

In Figure~\ref{fig:wordcloud_all}, we plot a word cloud of the inferred preferences across all conversation topics. Here we see the word cloud is dominated by generic preferences like ``conciseness'' and ``correct information,'' indicating a universal preference for precise and direct answers. This word cloud should be interpreted as the mean preferences of a generic person. These preferences generally tend towards accurate and clear responses. Additionally, the technical context some of the preferences indicate (like ``correct calculation) denote the technical topic bias in the data.

\begin{table}[t]
\begin{center}
\begin{tabular}{p{0.15\textwidth}p{0.1\textwidth}p{0.1\textwidth}p{0.25\textwidth}p{0.25\textwidth}}
\toprule
\textbf{Preference Category} & \textbf{\% of Data} & \textbf{\# of Preferences} & \textbf{Most prevalent in} & \textbf{Examples of granular preferences} \\
\midrule
\RaggedRight Clarity & \RaggedRight 48.22\% & 474 & \RaggedRight Computer Science / AI, Engineering and Technology & \RaggedRight Situational Awareness, Visual or Spatial Imagery \\
\midrule
\RaggedRight Thoroughness & \RaggedRight 39.16\% & 414 & \RaggedRight Politics, Agriculture / Food & \RaggedRight Detail, Compositional Depth \\
\midrule
\RaggedRight Accuracy & \RaggedRight 28.53\% & 248 & \RaggedRight Sports, History & \RaggedRight Precision, Accuracy in Context Application \\
\midrule
\RaggedRight Concise & \RaggedRight 15.32\% & 28 & \RaggedRight General Knowledge, Sports & \RaggedRight Simplified Explanation, Simplicity of Language \\
\midrule
\RaggedRight Relevance & \RaggedRight 15.13\% & 202 & \RaggedRight General Knowledge, Arts and Humanities & \RaggedRight Relevance to Query, Relevance and Accuracy \\
\midrule
\RaggedRight Engagement & \RaggedRight 11.15\% & 237 & \RaggedRight Writing and Literature, Creativity / Innovation & \RaggedRight Engagement and Enthusiasm, Engagement with Humor \\
\midrule
\RaggedRight Innovation & \RaggedRight 5.18\% & 96 & \RaggedRight Writing and Literature, Creativity / Innovation & \RaggedRight Originality, Interpretation of Creativity \\
\midrule
\RaggedRight Practicality & \RaggedRight 4.09\% & 73 & \RaggedRight Creativity / Innovation, Career and Personal Development & \RaggedRight Practicality of Suggestions, Practicality of Solution \\
\midrule
\RaggedRight Informative & \RaggedRight 4.08\% & 20 & \RaggedRight Natural Sciences, Sports & \RaggedRight Informative Details, Educational Approach \\
\midrule
\RaggedRight Diversity & \RaggedRight 3.16\% & 160 & \RaggedRight Agriculture / Food, Social Sciences & \RaggedRight Variety in Response Options, Acknowledgment of Diverse Perspectives \\
\midrule
\RaggedRight Comprehension & \RaggedRight 3.14\% & 119 & \RaggedRight Social Sciences, Psychology & \RaggedRight Empathy and Understanding in Approach, Insight \\
\bottomrule
\end{tabular}
\end{center}
\caption{First part of the final set of preferences (expanding on Table~\ref{tbl:preferences} to the top-11 preferences). Note that datapoints may have multiple preferences; the majority have two. So the '\% of Data' column does not sum to 100.}
\label{tbl:preferences_all_a}
\end{table}

\begin{table}[t]
\begin{center}
\begin{tabular}{p{0.15\textwidth}p{0.1\textwidth}p{0.1\textwidth}p{0.25\textwidth}p{0.25\textwidth}}
\toprule
\textbf{Preference Category} & \textbf{\% of Data} & \textbf{\# of Preferences} & \textbf{Most prevalent in} & \textbf{Examples of granular preferences} \\
\midrule
\RaggedRight Organization & \RaggedRight 2.92\% & 94 & \RaggedRight Education, Writing and Literature & \RaggedRight List Structure, Structure Detail \\
\midrule
\RaggedRight Follows Instructions & \RaggedRight 2.69\% & 165 & \RaggedRight Writing and Literature, Business & \RaggedRight Adherence to Requested Steps, Alignment with Given Data \\
\midrule
\RaggedRight Customization & \RaggedRight 1.71\% & 37 & \RaggedRight Agriculture / Food, Psychology & \RaggedRight Personalized Opinion, Personalized Advice \\
\midrule
\RaggedRight Concentration & \RaggedRight 1.68\% & 122 & \RaggedRight Creativity / Innovation, Politics & \RaggedRight Focus on Social Aspects, Focus \\
\midrule
\RaggedRight Helpfulness & \RaggedRight 1.65\% & 41 & \RaggedRight General Knowledge, Career and Personal Development & \RaggedRight Assistance Offering, Community Support \\
\midrule
\RaggedRight Humor & \RaggedRight 1.38\% & 15 & \RaggedRight Arts and Humanities, Culture and Society & \RaggedRight Humor Involvement, Humor and Wit \\
\midrule
\RaggedRight Context & \RaggedRight 1.32\% & 47 & \RaggedRight Culture and Society, Politics & \RaggedRight Contextual, Contextual Information \\
\midrule
\RaggedRight Environment & \RaggedRight 1.30\% & 30 & \RaggedRight Language and Communication, Writing and Literature & \RaggedRight Tone and Emotion, Tone and Reassurance \\
\midrule
\RaggedRight Direction & \RaggedRight 1.16\% & 92 & \RaggedRight Health, Psychology & \RaggedRight Decision-making, Guidance on Decision Making \\
\midrule
\RaggedRight Efficiency & \RaggedRight 1.13\% & 38 & \RaggedRight Writing and Literature, Psychology & \RaggedRight Performance and Efficiency, Potential Impact \\
\bottomrule
\end{tabular}
\end{center}
\caption{Second part of the final set of preferences (showing the bottom 10 preferences). Note that datapoints may have multiple preferences; the majority have two. So the '\% of Data' column does not sum to 100.}
\label{tbl:preferences_all_b}
\end{table}

\begin{table}[t]
\begin{center}
\begin{tabular}{p{0.3\textwidth}p{0.1\textwidth}p{0.1\textwidth}p{0.3\textwidth}}
\toprule
\textbf{Topic Category} & \textbf{\% of Data} & \textbf{\# of Topics} & \textbf{Most distinctive preferences categories} \\
\midrule
\RaggedRight Engineering and Technology & \RaggedRight 27.43\% & 304 & \RaggedRight Clarity, Accuracy \\
\midrule
\RaggedRight Arts and Humanities & \RaggedRight 17.48\% & 368 & \RaggedRight Humor, Innovation \\
\midrule
\RaggedRight Computer Science / AI & \RaggedRight 9.92\% & 179 & \RaggedRight Clarity, Direction \\
\midrule
\RaggedRight Business & \RaggedRight 6.42\% & 270 & \RaggedRight Environment, Follows Instructions \\
\midrule
\RaggedRight Social Sciences & \RaggedRight 4.73\% & 67 & \RaggedRight Customization, Diversity \\
\midrule
\RaggedRight Language and Communication & \RaggedRight 4.66\% & 185 & \RaggedRight Environment, Innovation \\
\midrule
\RaggedRight Health & \RaggedRight 3.50\% & 146 & \RaggedRight Direction, Helpfulness \\
\midrule
\RaggedRight Writing and Literature & \RaggedRight 3.46\% & 117 & \RaggedRight Efficiency, Innovation \\
\midrule
\RaggedRight Psychology & \RaggedRight 3.10\% & 125 & \RaggedRight Efficiency, Helpfulness \\
\midrule
\RaggedRight Philosophy & \RaggedRight 2.86\% & 107 & \RaggedRight Customization, Engagement \\
\midrule
\RaggedRight Career and Personal Development & \RaggedRight 2.65\% & 112 & \RaggedRight Helpfulness, Practicality \\
\midrule
\RaggedRight Education & \RaggedRight 2.61\% & 115 & \RaggedRight Helpfulness, Organization \\
\midrule
\RaggedRight History & \RaggedRight 2.28\% & 49 & \RaggedRight Informative, Accuracy \\
\midrule
\RaggedRight Politics & \RaggedRight 2.27\% & 60 & \RaggedRight Concentration, Informative \\
\midrule
\RaggedRight Natural Sciences & \RaggedRight 1.70\% & 96 & \RaggedRight Informative, Concentration \\
\midrule
\RaggedRight Culture and Society & \RaggedRight 1.15\% & 89 & \RaggedRight Humor, Context \\
\midrule
\RaggedRight Sports & \RaggedRight 1.13\% & 11 & \RaggedRight Informative, Accuracy \\
\midrule
\RaggedRight Leisure and Hobbies & \RaggedRight 0.86\% & 26 & \RaggedRight Direction, Diversity \\
\midrule
\RaggedRight General Knowledge & \RaggedRight 0.69\% & 34 & \RaggedRight Helpfulness, Customization \\
\midrule
\RaggedRight Creativity / Innovation & \RaggedRight 0.55\% & 10 & \RaggedRight Innovation, Concentration \\
\midrule
\RaggedRight Agriculture / Food & \RaggedRight 0.55\% & 23 & \RaggedRight Customization, Diversity \\
\bottomrule
\end{tabular}
\end{center}
\caption{The final set of topics (expanding on Table~\ref{tbl:topics}).}
\label{tbl:topics_all}
\end{table}

\section{Preference Distribution across Topics}\label{app:preference_distribution_across_topics}
\begin{figure}[h]
    \centering
    \begin{subfigure}{0.9\textwidth}
        \includegraphics[width=\linewidth]{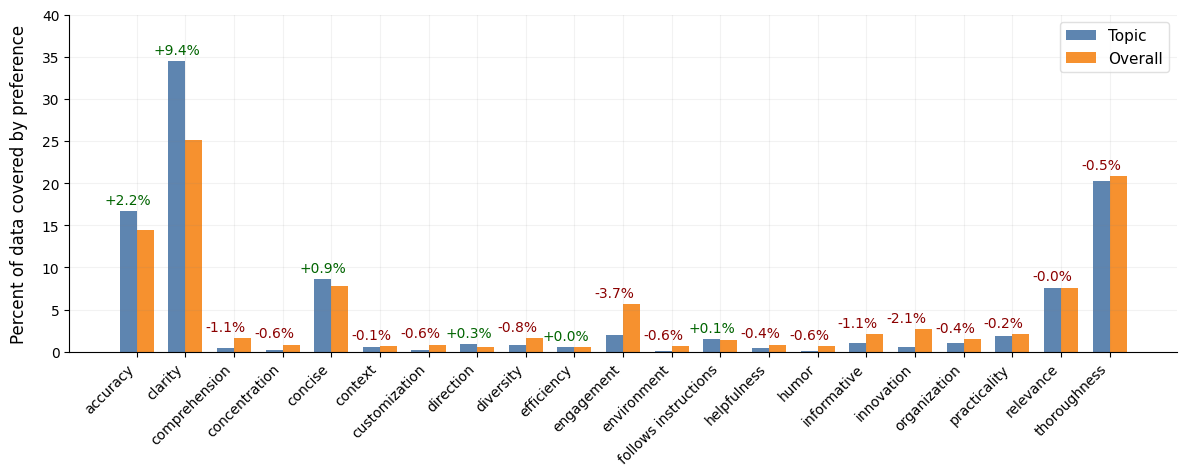}
        \caption{Computer Science and AI}
        \label{fig:preference_distribution_a}
    \end{subfigure}
    \par\bigskip
    \begin{subfigure}{0.9\textwidth}
        \includegraphics[width=\linewidth]{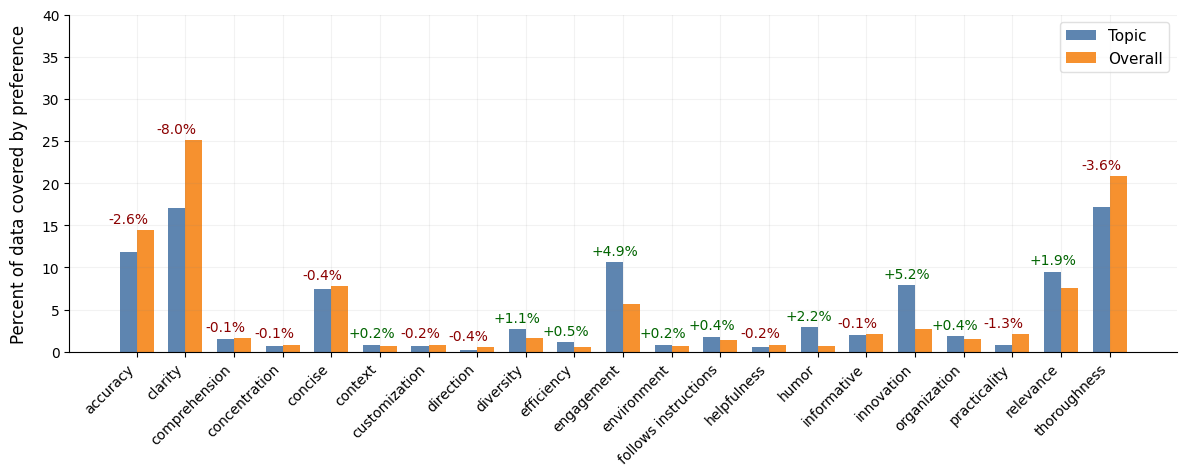} 
        \caption{Arts and Humanities}
        \label{fig:preference_distribution_b}
    \end{subfigure}
    \caption{Preference distribution across two topics. Y-axis shows that percent of data in the topic that the preference has been associated with. The X-axis shows the 21 preference categories. We show two bars: ``Topic'' refers to the preference distribution in the given topic; ``Overall'' refers to the overall (across all topics) preference distribution. 
    The numbers above each bar shows the delta from preference to baseline.}
    \label{fig:preference_distribution}
\end{figure}

In Figure~\ref{fig:preference_distribution}, we show the distribution of preferences across two topics: ``Computer Science and AI'' and ``Arts and Humanities.''
As described in Section~\ref{sec:human_preference_summaries}, the types of preferences users typically have varies significantly across topics.

\section{Additional Validation Results}\label{app:additional_results}
\begin{figure}[h]
    \centering
    \includegraphics[width=4.5in]{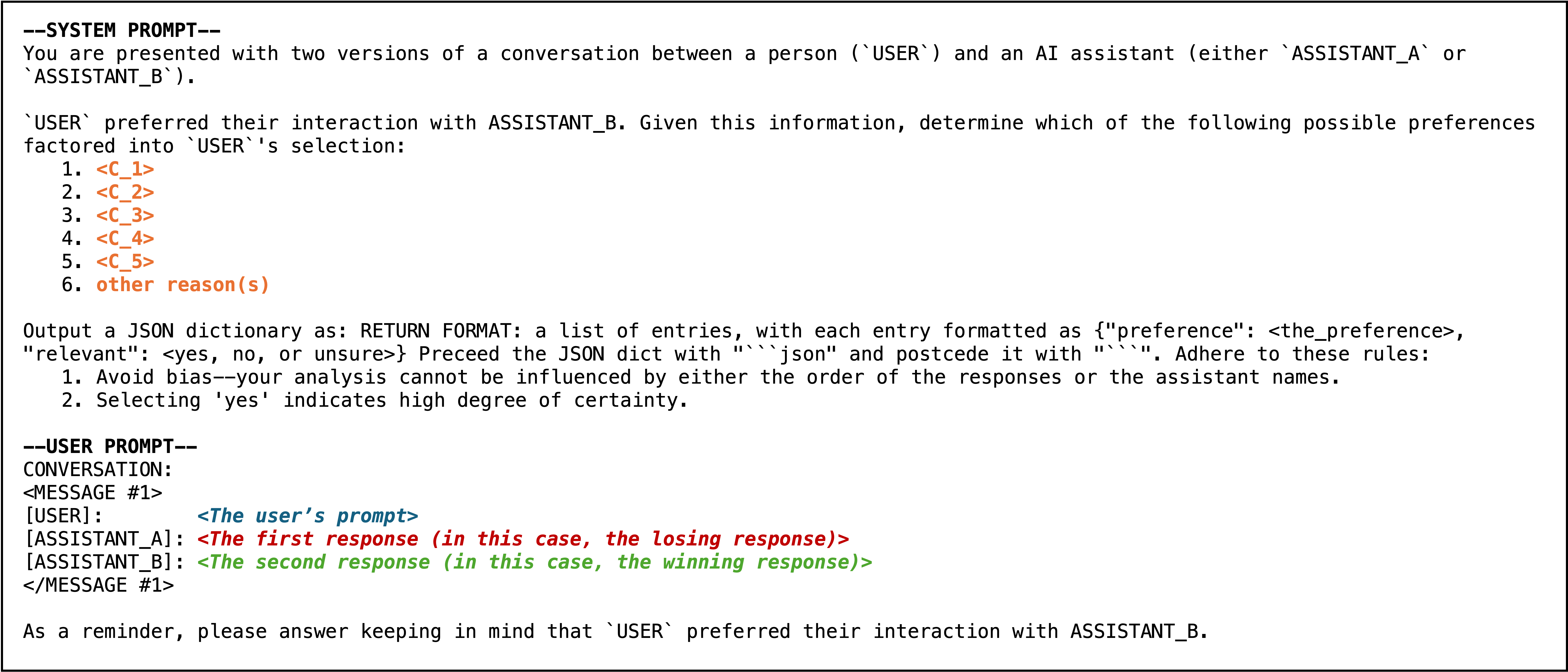}
    \caption{Prompt for evaluating preferences using LLMs.}
    \label{fig:synthetic_eval_prompt}
\end{figure}

In Figure~\ref{fig:synthetic_eval_prompt}, we provide the prompt for the synthetic evaluations to validate our preference groupings. Here, $C_i: i \in 1,\dots, 5$ are sampled from five subsets of fine-grained preferences as described in Section~\ref{subsec:eval_methods}. Note that the order of $C_i$ are randomized and the order of the assistant's responses (i.e., whether the selected response appears first or second) is also randomized. For each given conversation, the randomized order is kept identical across all models for the synthetic evaluation and across the human evaluators as well.

\begin{figure}[h]
    \centering
    \includegraphics[width=5.5in]{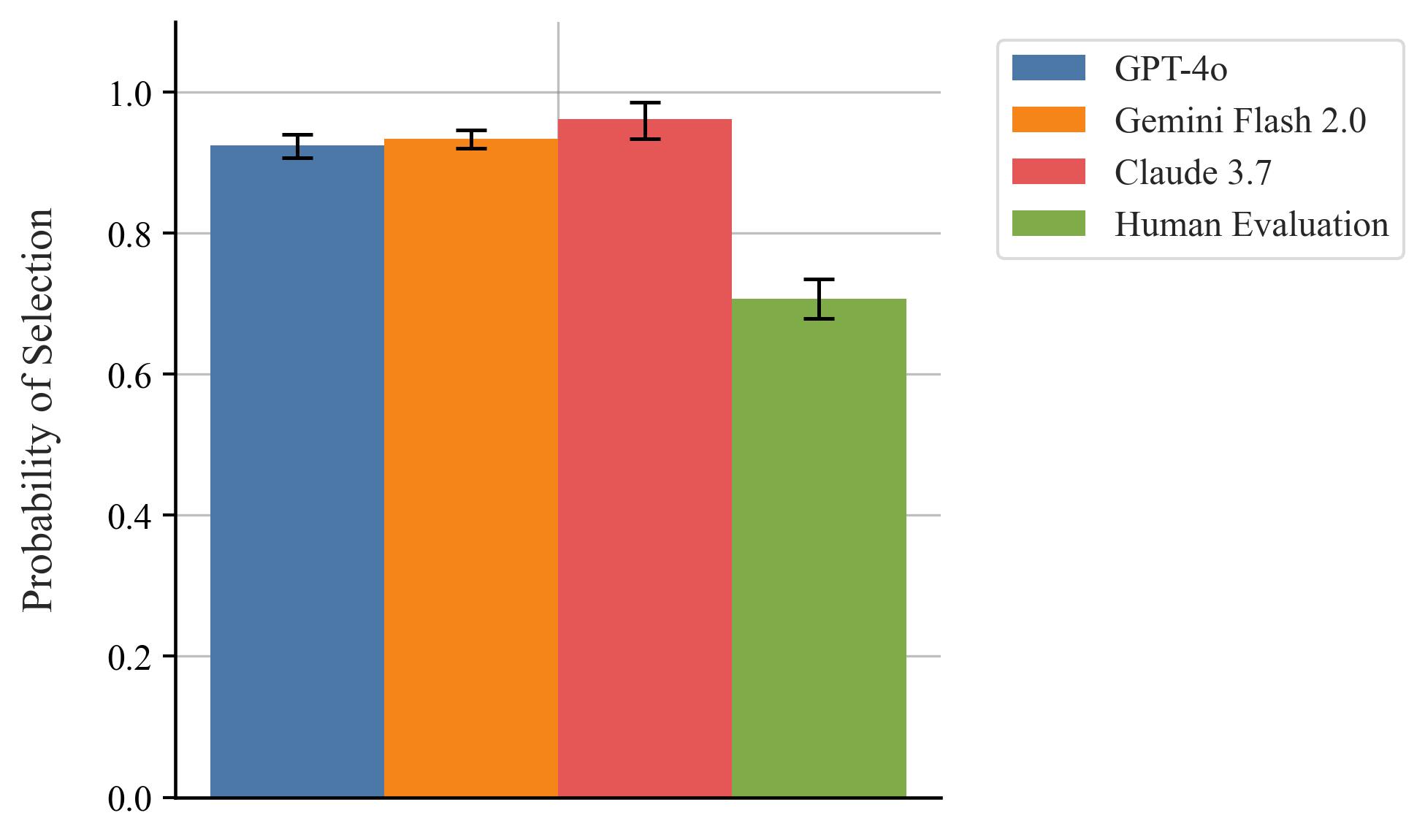}
    \caption{Probability of selecting the GPT-4o generated preference. Comparison using GPT-4o \cite{hurst2024gpt}, Gemini \cite{team2023gemini}, Claude 3.7, and human evaluations.}
    \label{fig:base_probabilities}
\end{figure}

In Figure~\ref{fig:base_probabilities}, we plot the absolute probability of selecting the generated preference across three LLMs and the human evaluations. This evaluation clearly confirms a strong bias towards the generated preference, indicating that the generated preferences are valid. Notably, while GPT-4o was used to generated these preferences, both Gemini and Claude models selected the preference at similar rates. The human selection rate is lower (about 70\% on average); however, this is likely due to high-levels of noise in human experiments. To confirm the human validation, we use the metrics described above.

\section{Human Evaluation Survey}\label{app:survey_ui}

\begin{figure}[htbp]
    \centering
    
    \includegraphics[width=0.6\textwidth]{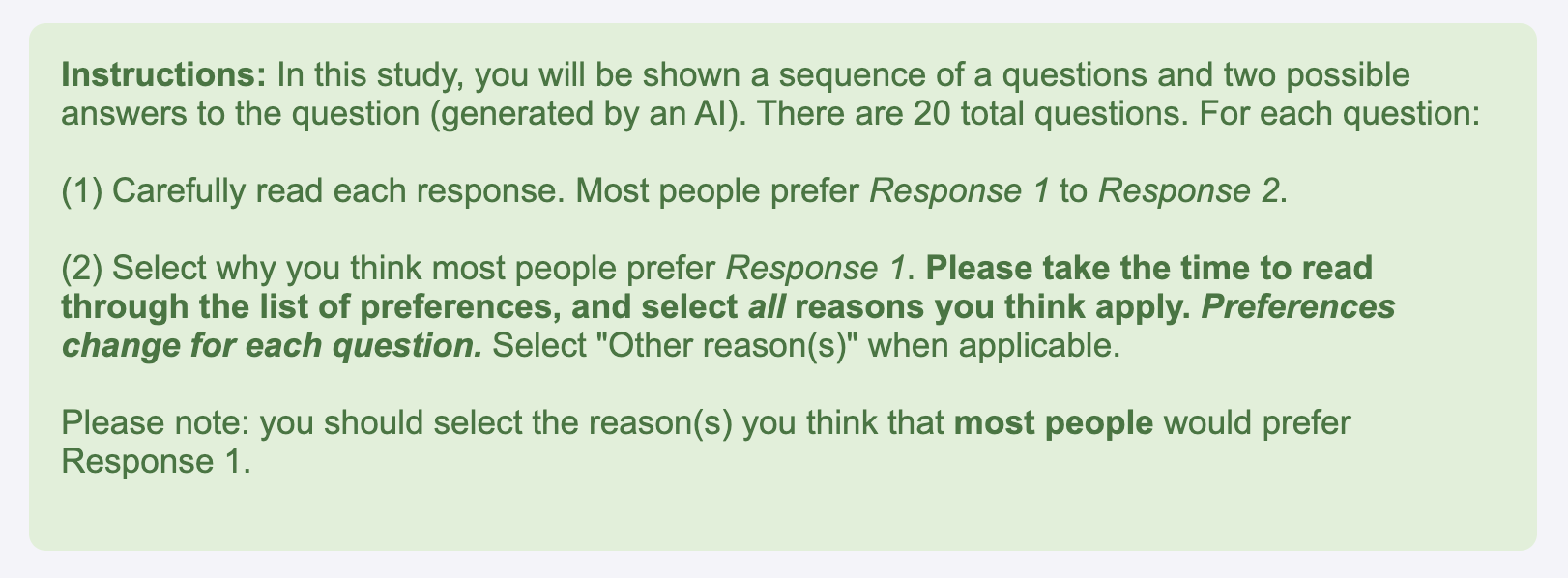}
    
    \subcaption{Instructions for the survey.}
    \label{fig:survey_ui_inst} 
    
    \vspace{1em} 
    
    \begin{subfigure}{0.45\textwidth}
        \centering
        \includegraphics[width=\textwidth]{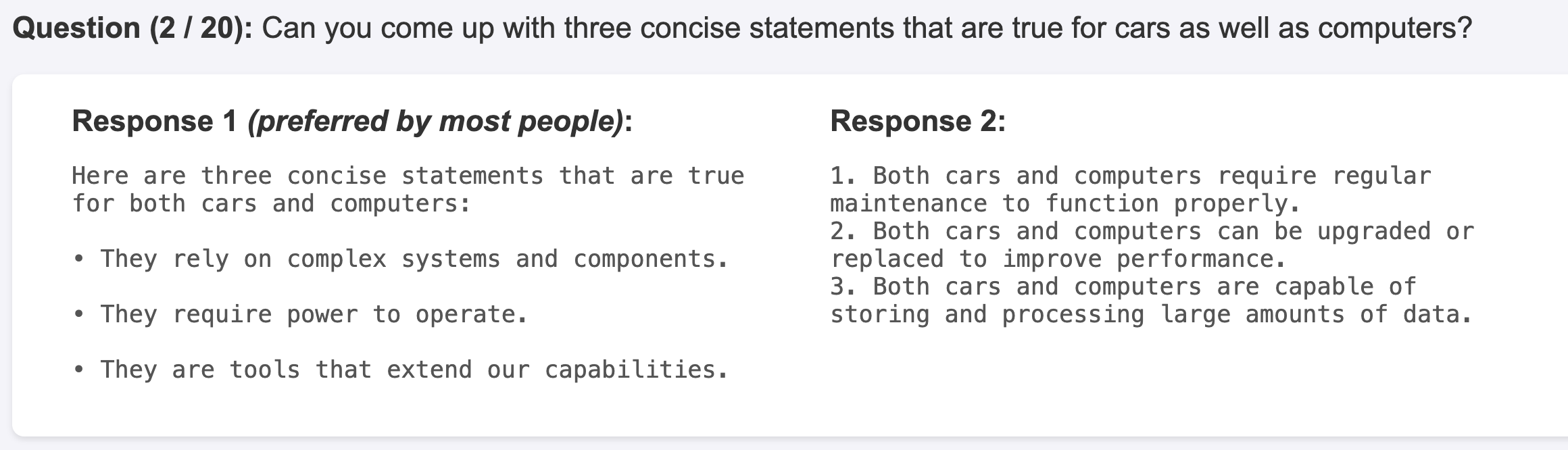}
        \subcaption{An example question showing two responses to a prompt. Only single-turn conversations were shown for simplicity.}
        \label{fig:survey_ui_q} 
    \end{subfigure}
    \hfill 
    \begin{subfigure}{0.45\textwidth}
        \centering
        \includegraphics[width=\textwidth]{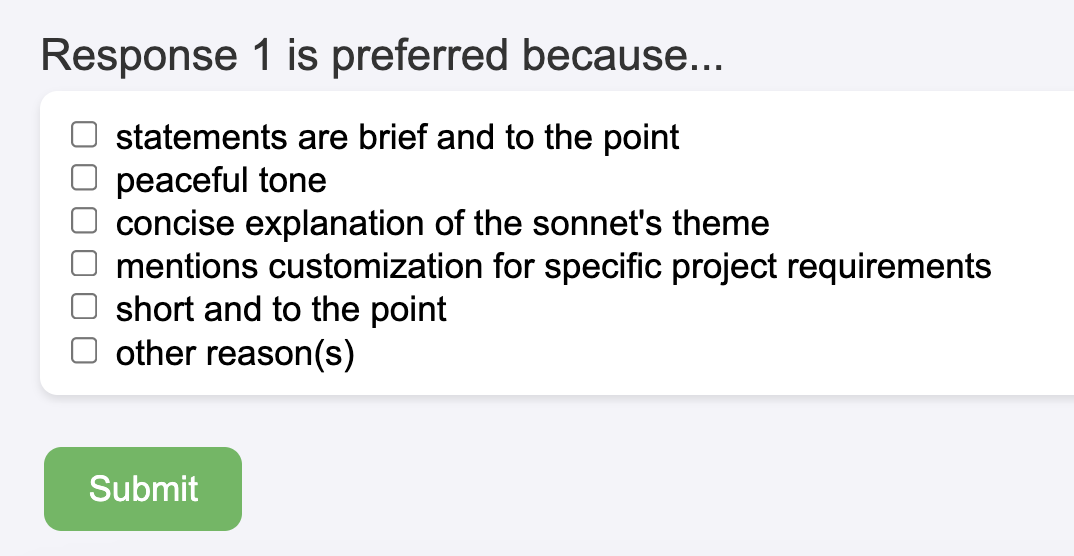}
        \subcaption{The responses available on the survey. Multiple options may be checked.}
        \label{fig:survey_ui_r} 
    \end{subfigure}

\end{figure}

Here we provide more information on the human validation survey discussed in Section~\ref{subsec:human_evaluation}. The survey layout is shown in Figures~\ref{fig:survey_ui_inst}, \ref{fig:survey_ui_q}, and \ref{fig:survey_ui_r}. Raters were sampled from cohort of English-speaking residents of the US. Each rater was paid at a recommended wage of \$12/hour.

\section{Additional details on Model Fine-tuning and Evaluation}\label{app:model_finetuning_and_eval}

\subsection{Preference Elo Results}
Preference Elo rankings are shown in Table~\ref{tbl:elo}. Here we show the overall ranking (this is what arena-style benchmarks show in their leaderboards) as well as preference-specific Elo rankings for four preferences: ``concise,'' ``humor,'' ``diversity,'' and ``concentration.'' GPT-4 is the overall winner, but not consistently so across all preferences. Variation in ranking is common and sometimes drastic across models. These results clearly indicate that different models are aligned to different preferences. Additionally, given our results in Figure~\ref{fig:preference_distribution} which indicate significant variation in user preference by topic (and so, we can also infer, by task), these Elo rankings suggest that different models are better aligned to human preferences on differing tasks.

\subsection{Fine-tuning details}
Here we include additional details on the model evaluation and fine-tuning setup discussed in Section~\ref{sec:applications_to_models}. We fine-tune a Qwen2 7B Instruct \cite{qwen2} and Ministral 8B Instruct \cite{mistralai_ministral8b_instruct_hf_2024} model using LoRA \cite{hu2022lora} with DPO applied to a preference-defined subset of our dataset. Half of the preference-defined subset is randomly held-out for testing. The hyperparameters are kept identical across all preferences and models. We add LoRA weights to the key, query, value, and output layers. We set $r:=128$ and $\alpha:=256$. We train for 2 epochs (regardless of the number of datapoints, which does vary across preferences) with an initial learning rate of $5\times10^{-6}$.

\subsection{Fine-tuning evaluation details}
We evaluate the fine-tuned models by adapting the LLM-As-A-Judge framework proposed in \cite{zheng2023judging}. In particular, we use a set of criteria (3 ``desired'' and 3 ``undesired'') to judge each preference. Criteria are generated synthetically through a few-shot prompt. The LLM judge assigns a binary value (0 or 1) for each criteria. The score is computed as the sum of desired criteria minus the sum of the undesired criteria. Because we use the scoring variant of the LLM judge, we run it independently for each prompt-response pair (i.e., we do not show multiple responses together). As a last step to make our results more robust, we run the judge three times for each response and keep the median score. 

\subsection{Fine-tuning results}
In Table~\ref{tbl:fine_tune_result}, we include information on fine-tuning performance across all preferences where we observed a significant change in performance. Over roughly $40\%$ of preference categories (and across models), we observe a significant increase in performance. The reason this is not higher is likely for three reasons. This mode of fine-tuning is likely not as suitable for some preferences like ``accuracy,'' as it is known that LoRA fine-tuning is not consistently effective for learning factual information, and rather is more suitable for adapting to stylistic preferences \cite{ratnakar2025beyond}. Additionally, the number of examples is relatively small for some preferences (recall that $50\%$ of datapoints are held-out as test data. And finally, we do not adjust hyperparameters across preference or model settings, so these hyperparameters are likely suboptimal. And despite these limitations, we still observe significant improvement on close to half of the preferences.

\begin{figure}[h]
    \centering
    \includegraphics[width=5.5in]{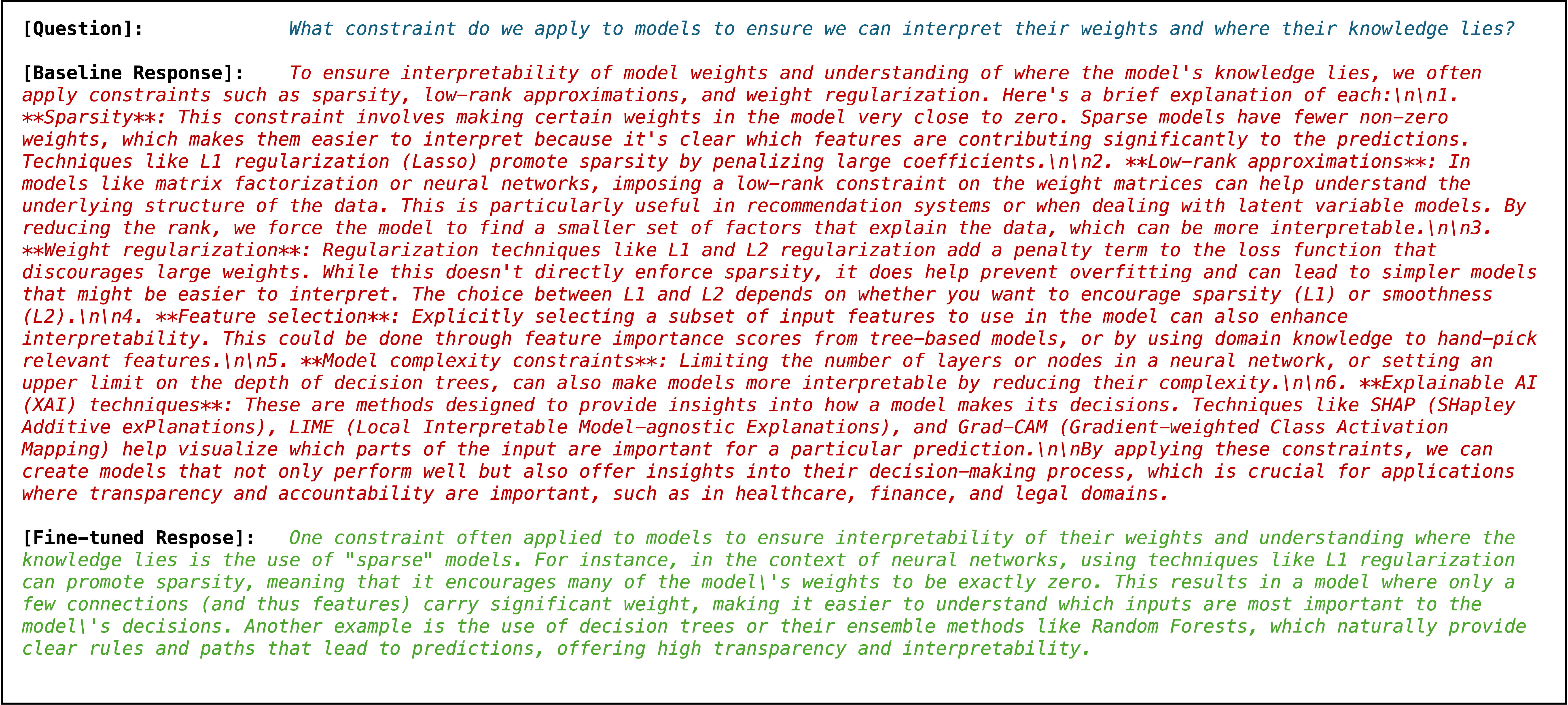}
    \caption{Example showing a user's question and the response of Qwen2 7B Instruct before and after fine-tuning on the ``concise-preference'' subset of Chatbot Arena data. This example comes from a held-out test set. Across all test data, fine-tuning reduces the average response length by about $60\%$.}
    \label{fig:concise_finetune_example}
\end{figure}

These improvements are sometimes qualitatively obvious. For example, as mentioned in Section~\ref{sec:applications_to_models}, fine-tuning on the ``concise'' subset leads to a $60\%$ reduction in response length. An example is presented in Figure~\ref{fig:concise_finetune_example} to show how this works qualitatively.

\begin{table}[t]
\begin{center}
\begin{tabular}{p{0.15\textwidth}p{0.1\textwidth}p{0.1\textwidth}p{0.1\textwidth}p{0.1\textwidth}p{0.1\textwidth}p{0.1\textwidth}}
\toprule
\textbf{Preference} & \textbf{1st} & \textbf{2nd} & \textbf{3rd} & \textbf{4th} & \textbf{5th} & \textbf{6th} \\
\midrule
\RaggedRight Overall & \RaggedRight GPT-4 & \RaggedRight Claude 3.5 Sonnet & \RaggedRight Claude 3.5 Haiku & \RaggedRight GPT-3.5 Turbo & \RaggedRight Guanaco 33B & \RaggedRight Palm 2 \\
\midrule
\RaggedRight Concise & \RaggedRight GPT-3.5 Turbo & \RaggedRight Alpaca 13B & \RaggedRight GPT-4 & \RaggedRight Claude 3.5 Sonnet & \RaggedRight Claude 3.5 Haiku & \RaggedRight MPT 7B Chat \\
\midrule
\RaggedRight Humor & \RaggedRight GPT-4 & \RaggedRight Claude 3.5 Haiku & \RaggedRight GPT-3.5 Turbo & \RaggedRight Claude 3.5 Sonnet & \RaggedRight Palm 2 & \RaggedRight Guanaco 33B \\
\midrule
\RaggedRight Diversity & \RaggedRight Claude 3.5 Sonnet & \RaggedRight Palm 2 & \RaggedRight Claude 3.5 Haiku & \RaggedRight GPT-4 & \RaggedRight GPT-3.5 Turbo & \RaggedRight Vicuna 13B \\
\midrule
\RaggedRight Concentration & \RaggedRight Alpaca 13B & \RaggedRight GPT-3.5 Turbo & \RaggedRight GPT-4 & \RaggedRight WizardLM 13B & \RaggedRight Guanaco 33B & \RaggedRight RWKV-4 Raven 14B \\
\bottomrule
\end{tabular}
\end{center}
\caption{Preference-specific Elo Rankings taken from Chatbot Arena dataset. Rankings are computed across preference-defined subsets. Showing the top-6 models for a sample for preferences.}
\label{tbl:elo}
\end{table}

\begin{table}[t]
\begin{center}
\begin{tabular}{p{0.25\textwidth}p{0.25\textwidth}p{0.3\textwidth}}
\toprule
\textbf{Model} & \textbf{Preference Category} & \textbf{LLM Judge Score Difference} \\
\midrule
\multirow{6}{*}{\RaggedRight Qwen2 (7B, Instruct)} & \RaggedRight \textbf{Concise} & \RaggedRight 0.59 +/- (0.83, 0.33) \\
\cmidrule(lr){2-3}
 & \RaggedRight \textbf{Context} & \RaggedRight 0.22 +/- (0.41, 0.03) \\
\cmidrule(lr){2-3}
 & \RaggedRight \textbf{Engagement} & \RaggedRight 0.53 +/- (0.74, 0.33) \\
\cmidrule(lr){2-3}
 & \RaggedRight Follows Instructions & \RaggedRight -0.21 +/- (-0.05, -0.36) \\
\cmidrule(lr){2-3}
 & \RaggedRight \textbf{Humor} & \RaggedRight 0.44 +/- (0.68, 0.21) \\
\cmidrule(lr){2-3}
 & \RaggedRight \textbf{Innovation} & \RaggedRight 0.64 +/- (0.95, 0.34) \\
\midrule
\multirow{11}{*}{\RaggedRight Ministral (8B, Instruct)} & \RaggedRight Clarity & \RaggedRight -0.22 +/- (-0.11, -0.33) \\
\cmidrule(lr){2-3}
 & \RaggedRight \textbf{Concise} & \RaggedRight 0.74 +/- (0.91, 0.56) \\
\cmidrule(lr){2-3}
 & \RaggedRight \textbf{Customization} & \RaggedRight 0.30 +/- (0.54, 0.07) \\
\cmidrule(lr){2-3}
 & \RaggedRight Direction & \RaggedRight -0.31 +/- (-0.19, -0.43) \\
\cmidrule(lr){2-3}
 & \RaggedRight \textbf{Diversity} & \RaggedRight 0.55 +/- (0.73, 0.37) \\
\cmidrule(lr){2-3}
 & \RaggedRight \textbf{Efficiency} & \RaggedRight 0.14 +/- (0.25, 0.04) \\
\cmidrule(lr){2-3}
 & \RaggedRight \textbf{Engagement} & \RaggedRight 0.58 +/- (0.76, 0.41) \\
\cmidrule(lr){2-3}
 & \RaggedRight \textbf{Environment} & \RaggedRight 0.16 +/- (0.28, 0.05) \\
\cmidrule(lr){2-3}
 & \RaggedRight \textbf{Helpfulness} & \RaggedRight 0.12 +/- (0.23, 0.03) \\
\cmidrule(lr){2-3}
 & \RaggedRight \textbf{Humor} & \RaggedRight 0.33 +/- (0.59, 0.06) \\
\cmidrule(lr){2-3}
 & \RaggedRight \textbf{Practicality} & \RaggedRight 0.21 +/- (0.35, 0.07) \\
\bottomrule
\end{tabular}
\end{center}
\caption{Fine-tuning on preference-defined subsets leads to significant improvement in model performance on that preference. Bolded values for positive change (i.e., fine-tuning improves performance.)}
\label{tbl:fine_tune_result}
\end{table}

%% file: main.bbl
\begin{thebibliography}{10}

\bibitem{amini2024direct}
Afra Amini, Tim Vieira, and Ryan Cotterell.
\newblock Direct preference optimization with an offset.
\newblock {\em arXiv preprint arXiv:2402.10571}, 2024.

\bibitem{anthropic_claude_3_7_sonnet_2025}
Anthropic.
\newblock Claude 3.7 sonnet and claude code.
\newblock Anthropic Blog, February 2025.

\bibitem{azar2024general}
Mohammad~Gheshlaghi Azar, Zhaohan~Daniel Guo, Bilal Piot, Remi Munos, Mark Rowland, Michal Valko, and Daniele Calandriello.
\newblock A general theoretical paradigm to understand learning from human preferences.
\newblock In {\em International Conference on Artificial Intelligence and Statistics}, pages 4447--4455. PMLR, 2024.

\bibitem{bai2022training}
Yuntao Bai, Andy Jones, Kamal Ndousse, Amanda Askell, Anna Chen, Nova DasSarma, Dawn Drain, Stanislav Fort, Deep Ganguli, Tom Henighan, Nicholas Joseph, Saurav Kadavath, Jackson Kernion, Tom Conerly, Sheer~El Showk, Nelson Elhage, Zac Hatfield-Dodds, Danny Hernandez, Tristan Hume, Scott Johnston, Shauna Kravec, Liane Lovitt, Neel Nanda, Catherine Olsson, Dario Amodei, Tom Brown, Jack Clark, Sam McCandlish, Chris Olah, Ben Mann, and Jared Kaplan.
\newblock Training a helpful and harmless assistant with reinforcement learning from human feedback, 2022.

\bibitem{bai2022constitutional}
Yuntao Bai, Saurav Kadavath, Sandipan Kundu, Amanda Askell, Jackson Kernion, Andy Jones, Anna Chen, Anna Goldie, Azalia Mirhoseini, Cameron McKinnon, et~al.
\newblock Constitutional ai: Harmlessness from ai feedback.
\newblock {\em arXiv preprint arXiv:2212.08073}, 2022.

\bibitem{chiang2024chatbot}
Wei-Lin Chiang, Lianmin Zheng, Ying Sheng, Anastasios~Nikolas Angelopoulos, Tianle Li, Dacheng Li, Hao Zhang, Banghua Zhu, Michael Jordan, Joseph~E Gonzalez, et~al.
\newblock Chatbot arena: An open platform for evaluating llms by human preference.
\newblock {\em arXiv preprint arXiv:2403.04132}, 2024.

\bibitem{christiano2017deep}
Paul~F Christiano, Jan Leike, Tom Brown, Miljan Martic, Shane Legg, and Dario Amodei.
\newblock Deep reinforcement learning from human preferences.
\newblock {\em Advances in neural information processing systems}, 30, 2017.

\bibitem{costa1999five}
PT~Costa and RR~McCrae.
\newblock A five-factor theory of personality.
\newblock {\em Handbook of personality: Theory and research}, 2(01):1999, 1999.

\bibitem{pmlr-v162-ethayarajh22a}
Kawin Ethayarajh, Yejin Choi, and Swabha Swayamdipta.
\newblock Understanding dataset difficulty with $\mathcal{V}$-usable information.
\newblock In Kamalika Chaudhuri, Stefanie Jegelka, Le~Song, Csaba Szepesvari, Gang Niu, and Sivan Sabato, editors, {\em Proceedings of the 39th International Conference on Machine Learning}, volume 162 of {\em Proceedings of Machine Learning Research}, pages 5988--6008. PMLR, 17--23 Jul 2022.

\bibitem{ethayarajh2024kto}
Kawin Ethayarajh, Winnie Xu, Niklas Muennighoff, Dan Jurafsky, and Douwe Kiela.
\newblock Kto: Model alignment as prospect theoretic optimization.
\newblock {\em arXiv preprint arXiv:2402.01306}, 2024.

\bibitem{hathaway1951minnesota}
Starke~Rosecrans Hathaway and John~Charnley McKinley.
\newblock Minnesota multiphasic personality inventory; manual, revised.
\newblock 1951.

\bibitem{hu2022lora}
Edward~J Hu, Yelong Shen, Phillip Wallis, Zeyuan Allen-Zhu, Yuanzhi Li, Shean Wang, Lu~Wang, Weizhu Chen, et~al.
\newblock Lora: Low-rank adaptation of large language models.
\newblock {\em ICLR}, 1(2):3, 2022.

\bibitem{huang2024text}
Chen Huang and Guoxiu He.
\newblock Text clustering as classification with llms.
\newblock {\em arXiv preprint arXiv:2410.00927}, 2024.

\bibitem{hurst2024gpt}
Aaron Hurst, Adam Lerer, Adam~P Goucher, Adam Perelman, Aditya Ramesh, Aidan Clark, AJ~Ostrow, Akila Welihinda, Alan Hayes, Alec Radford, et~al.
\newblock Gpt-4o system card.
\newblock {\em arXiv preprint arXiv:2410.21276}, 2024.

\bibitem{kwon2023image}
Sehyun Kwon, Jaeseung Park, Minkyu Kim, Jaewoong Cho, Ernest~K Ryu, and Kangwook Lee.
\newblock Image clustering conditioned on text criteria.
\newblock {\em arXiv preprint arXiv:2310.18297}, 2023.

\bibitem{h4stackexchange}
Nathan Lambert, Lewis Tunstall, Nazneen Rajani, and Tristan Thrush.
\newblock Huggingface h4 stack exchange preference dataset, 2023.

\bibitem{lee2024mechanistic}
Andrew Lee, Xiaoyan Bai, Itamar Pres, Martin Wattenberg, Jonathan~K Kummerfeld, and Rada Mihalcea.
\newblock A mechanistic understanding of alignment algorithms: A case study on dpo and toxicity.
\newblock {\em arXiv preprint arXiv:2401.01967}, 2024.

\bibitem{mistralai_ministral8b_instruct_hf_2024}
{Mistral AI}.
\newblock Ministral-8b-instruct-2410.
\newblock Model Repository, 2024.

\bibitem{myers1962myers}
IB~Myers.
\newblock The myers-briggs type indicator.
\newblock {\em Educational Testing Service/Princeton}, 1962.

\bibitem{poddar2024personalizing}
Sriyash Poddar, Yanming Wan, Hamish Ivison, Abhishek Gupta, and Natasha Jaques.
\newblock Personalizing reinforcement learning from human feedback with variational preference learning.
\newblock {\em arXiv preprint arXiv:2408.10075}, 2024.

\bibitem{prolific}
\url{https://www.prolific.co}, 2024.

\bibitem{rafailov2024direct}
Rafael Rafailov, Archit Sharma, Eric Mitchell, Christopher~D Manning, Stefano Ermon, and Chelsea Finn.
\newblock Direct preference optimization: Your language model is secretly a reward model.
\newblock {\em Advances in Neural Information Processing Systems}, 36, 2024.

\bibitem{ratnakar2025beyond}
Shivam Ratnakar, Abhiroop Talasila, Raghav Chamadiya, Nikhil Agarwal, and Vinayak~K Doifode.
\newblock Beyond qa pairs: Assessing parameter-efficient fine-tuning for fact embedding in llms.
\newblock {\em arXiv preprint arXiv:2503.01131}, 2025.

\bibitem{salemi2023lamp}
Alireza Salemi, Sheshera Mysore, Michael Bendersky, and Hamed Zamani.
\newblock Lamp: When large language models meet personalization.
\newblock {\em arXiv preprint arXiv:2304.11406}, 2023.

\bibitem{singhal2023long}
Prasann Singhal, Tanya Goyal, Jiacheng Xu, and Greg Durrett.
\newblock A long way to go: Investigating length correlations in rlhf.
\newblock {\em arXiv preprint arXiv:2310.03716}, 2023.

\bibitem{tan2024democratizing}
Zhaoxuan Tan, Qingkai Zeng, Yijun Tian, Zheyuan Liu, Bing Yin, and Meng Jiang.
\newblock Democratizing large language models via personalized parameter-efficient fine-tuning.
\newblock {\em arXiv preprint arXiv:2402.04401}, 2024.

\bibitem{team2023gemini}
Gemini Team, Rohan Anil, Sebastian Borgeaud, Yonghui Wu, Jean-Baptiste Alayrac, Jiahui Yu, Radu Soricut, Johan Schalkwyk, Andrew~M Dai, Anja Hauth, et~al.
\newblock Gemini: a family of highly capable multimodal models.
\newblock {\em arXiv preprint arXiv:2312.11805}, 2023.

\bibitem{turk1991eigenfaces}
Matthew Turk and Alex Pentland.
\newblock Eigenfaces for recognition.
\newblock {\em Journal of cognitive neuroscience}, 3(1):71--86, 1991.

\bibitem{wang2024interpretable}
Haoxiang Wang, Wei Xiong, Tengyang Xie, Han Zhao, and Tong Zhang.
\newblock Interpretable preferences via multi-objective reward modeling and mixture-of-experts.
\newblock {\em arXiv preprint arXiv:2406.12845}, 2024.

\bibitem{wang2023helpsteer}
Zhilin Wang, Yi~Dong, Jiaqi Zeng, Virginia Adams, Makesh~Narsimhan Sreedhar, Daniel Egert, Olivier Delalleau, Jane~Polak Scowcroft, Neel Kant, Aidan Swope, et~al.
\newblock Helpsteer: Multi-attribute helpfulness dataset for steerlm.
\newblock {\em arXiv preprint arXiv:2311.09528}, 2023.

\bibitem{qwen2}
An~Yang, Baosong Yang, Binyuan Hui, Bo~Zheng, Bowen Yu, Chang Zhou, Chengpeng Li, Chengyuan Li, Dayiheng Liu, Fei Huang, Guanting Dong, Haoran Wei, Huan Lin, Jialong Tang, Jialin Wang, Jian Yang, Jianhong Tu, Jianwei Zhang, Jianxin Ma, Jin Xu, Jingren Zhou, Jinze Bai, Jinzheng He, Junyang Lin, Kai Dang, Keming Lu, Keqin Chen, Kexin Yang, Mei Li, Mingfeng Xue, Na~Ni, Pei Zhang, Peng Wang, Ru~Peng, Rui Men, Ruize Gao, Runji Lin, Shijie Wang, Shuai Bai, Sinan Tan, Tianhang Zhu, Tianhao Li, Tianyu Liu, Wenbin Ge, Xiaodong Deng, Xiaohuan Zhou, Xingzhang Ren, Xinyu Zhang, Xipin Wei, Xuancheng Ren, Yang Fan, Yang Yao, Yichang Zhang, Yu~Wan, Yunfei Chu, Yuqiong Liu, Zeyu Cui, Zhenru Zhang, and Zhihao Fan.
\newblock Qwen2 technical report.
\newblock {\em arXiv preprint arXiv:2407.10671}, 2024.

\bibitem{yang2024rewards}
Rui Yang, Xiaoman Pan, Feng Luo, Shuang Qiu, Han Zhong, Dong Yu, and Jianshu Chen.
\newblock Rewards-in-context: Multi-objective alignment of foundation models with dynamic preference adjustment.
\newblock {\em arXiv preprint arXiv:2402.10207}, 2024.

\bibitem{zheng2023judging}
Lianmin Zheng, Wei-Lin Chiang, Ying Sheng, Siyuan Zhuang, Zhanghao Wu, Yonghao Zhuang, Zi~Lin, Zhuohan Li, Dacheng Li, Eric.~P Xing, Hao Zhang, Joseph~E. Gonzalez, and Ion Stoica.
\newblock Judging llm-as-a-judge with mt-bench and chatbot arena, 2023.

\end{thebibliography}
